\documentclass{article}

\usepackage{arxiv}

\usepackage{amsmath}
\usepackage[utf8]{inputenc} % allow utf-8 input
\usepackage[T1]{fontenc}    % use 8-bit T1 fonts
\usepackage{hyperref}       % hyperlinks
\usepackage{url}            % simple URL typesetting
\usepackage{booktabs}       % professional-quality tables
\usepackage{amsfonts}       % blackboard math symbols
\usepackage{nicefrac}       % compact symbols for 1/2, etc.
\usepackage{microtype}      % microtypography
\usepackage{lipsum}
\usepackage{natbib}
\usepackage{graphicx}
\graphicspath{ {./images/} }

\usepackage{xcolor}         % colors
\usepackage[most]{tcolorbox}
\tcbuselibrary{skins,breakable}
\usepackage{fvextra}     % Verbatim + \VerbatimInput with wrapping
\usepackage{caption}     % for \captionof

\usepackage{tabularx}
\usepackage{array}
\usepackage{siunitx}
\usepackage{pifont}
\usepackage{makecell}
\usepackage{threeparttable}

\newcolumntype{Y}{>{\centering\arraybackslash}X}

\fvset{
  breaklines, breakanywhere,
  breaksymbol={}, breaksymbolleft={}, breaksymbolright={}, breaksymbolindent=0pt
}

\definecolor{promptbg}{HTML}{FAFAFA}
\definecolor{promptframe}{HTML}{D0D7DE}
\definecolor{systemc}{HTML}{6B7280}
\definecolor{userc}{HTML}{0B5394}

\tcbset{
  promptbox/.style={
    enhanced,
    colback=promptbg, colframe=promptframe,
    left=1em, right=1em, top=.2em, bottom=.2em,
    boxrule=.2pt, arc=2mm, before skip=1ex, after skip=1ex,
    coltitle=white
  }
}

\title{On the Strengths and Weaknesses of Data for Open-set Embodied Assistance}

\author{
 Pradyumna Tambwekar\thanks{Work completed while associated with Toyota Resesrch Institute}$^*$ \\
  Distyl AI\\
  \texttt{pradyumna.tambwekar@distyl.ai} \\
   \And
 Andrew Silva \\
  Toyota Research Institute\\
  \texttt{andrew.silva@tri.global} \\
  \And
 Deepak Gopinath  \\
  Toyota Research Institute\\
  \texttt{deepak.gopinath@tri.global} \\
  \And
Jonathan DeCastro  \\
  Toyota Research Institute\\
  \texttt{jonathan.decastro@tri.global} \\
\And
Xiongyi Cui  \\
  Toyota Research Institute\\
  \texttt{xiongyi.cui@tri.global} \\
\And
Guy Rosman  \\
  Toyota Research Institute\\
  \texttt{guy.rosman@tri.global} \\
}

\begin{document}
\maketitle
\begin{abstract}
Embodied foundation models are increasingly performant in real-world domains such as robotics or autonomous driving. These models are often deployed
in interactive or assistive settings, where it is important that these assistive models generalize to new users and new tasks. Diverse interactive data generation offers a promising avenue for providing data-efficient generalization capabilities for interactive embodied foundation models. 
In this paper, we investigate the generalization capabilities of a multimodal foundation model fine-tuned on diverse interactive assistance data in a synthetic domain. We explore generalization along two axes: a) assistance with unseen categories of user behavior and b) providing guidance in new configurations not encountered during training. 
We study a broad capability called \textbf{Open-Set Corrective Assistance}, in which the model needs to inspect lengthy user behavior and provide assistance through either corrective actions or language-based feedback. 
This task remains unsolved in prior work, which typically assumes closed corrective categories or relies on external planners, making it a challenging testbed for evaluating the limits of assistive data.
To support this task, we generate synthetic assistive datasets in Overcooked and fine-tune a LLaMA-based model to evaluate generalization to novel tasks and user behaviors. 
Our approach provides key insights into the nature of assistive datasets required to enable open-set assistive intelligence. In particular, we show that performant models benefit from datasets that cover different aspects of assistance, including multimodal grounding, defect inference, and exposure to diverse scenarios.
\end{abstract}

% keywords can be removed

\section{Introduction}

Embodied foundation models, once trained, are shown to generalize to new tasks or modalities with minimal additional task-specific data~\cite{zheng2025universal, 11098567, gao2024survey}. 
However, training schemes for these models often involve vast amounts of complex multimodal data, which are expensive and laborious to collect~\cite{DEMELO2022174}. 
Therefore, many recent approaches leverage intelligent synthetic data-generation/augmentation paradigms to increase the diversity of the tasks and data that the model gets exposed to during training.~\cite{ahn2024autort}. 
In recent work, foundational robotics models trained on large-scale synthetic datasets have demonstrated competitive generalization performance on tasks such as robotic manipulation~\cite{deng2025graspvla}. 
To transform embodied foundation models into fielded artifacts that interact with the real world, understanding and developing novel methods and modalities of synthetic data is a critical next step~\cite{ren2025cosmos}. 

This paper focuses specifically on embodied models which are aimed at assistive collaboration, i.e. \textit{interactive} foundation models. 
Interactive foundation models must be able to inspect and understand temporally-extended user-behavior via a multimodal history of the user's interactions with the assistive agent and environment~\cite{sinha2024real, liu2024heirarchical}. 
We focus on an unsolved task which we denote as Open-Set Corrective Assistance, wherein an assistive model needs to inspect user-behavior and provide assistance to correct any behavior that is counterproductive to the task-goals, without a prespecified finite set of tasks nor a finite set of counterproductive defects in behavior. 
Large-scale, real-world data collection in these settings remains challenging, as long-form interaction data in embodied environments is inherently noisy and often requires substantial monitoring and scaffolding~\cite{sumner2025simcoachcorpus}.
Furthermore, models trained in these settings need to generalize along multiple dimensions, i.e. differing behavior-types as well as differing tasks or domains~\cite{sui2025grounding}. 
Collectively, these constraints motivate diverse assistive data as an important tool for evaluating the scaling of interactive foundation models in embodied environments. 

\begin{figure}[tbp]
    \centering
    \includegraphics[width=\textwidth]{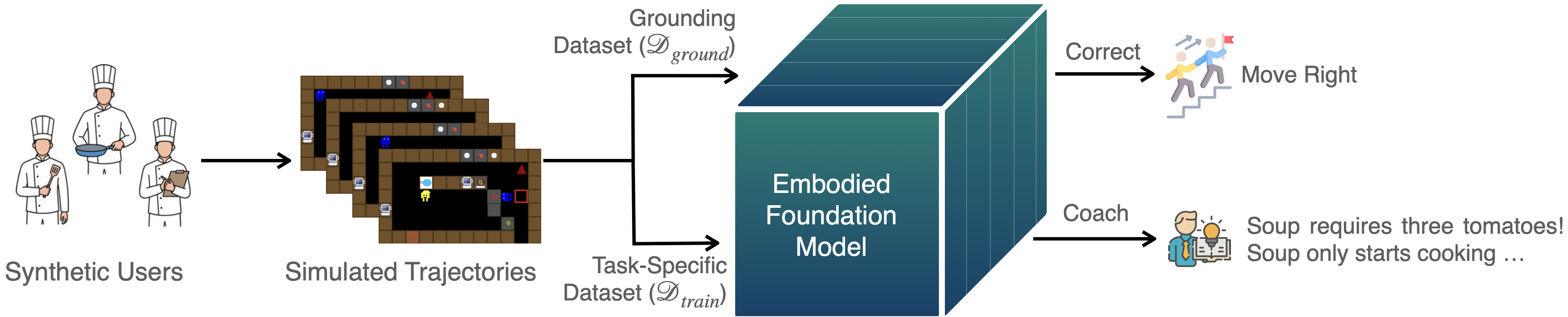}
    \caption{We simulate synthetic users in Overcooked to generate multimodal (image + text) gameplay trajectories, which used to  distill complementary synthetic datasets. These datasets are designed to (1) ground actions to environmental outcomes and (2) support behavior understanding and correction from trajectories. By training an embodied model on these data, we evaluate whether an embodied foundation model can generalize to unseen defective behaviors and novel task configurations.}
    \label{fig:overview}
\end{figure}

In this paper, we focus on testing the capabilities of a foundation model, trained on assistive data, so as to probe the influence of diverse embodied data on open-set assistive models. 
Recent work on assistive modeling has provided methods to generate corrective actions which a user can employ to remediate their suboptimal behavior~\cite{verghese2025user, gopinath2024computational}, or provide language-based coaching to enable the user to understand and correct any deficiencies~\cite{liu2023reflect, guan2024task}. 
However, these methods are limited by the requirement of a closed-set of possible corrections to select from, or are dependent on a domain-specific planner to execute high-level corrective strategies. 
By exploring this problem in an open-set scenario (without the assumption of a fixed set of corrections), we directly probe whether foundation models trained solely on a limited task data can generalize their domain knowledge towards previously unseen behaviors and task configurations.

In this paper, we instruction-tune~\cite{ouyang2022training} an assistive model capable of generating \textit{either} corrective actions or open-ended feedback to correct a deficiency in a user's behavior, entirely in natural language. 
Our model-architecture leverages Llama-3 suite models as the base, equipped with a ViT encoder to embed the image-based state representations across a trajectory into the embedding-space of the multimodal language model (MLM)~\cite{liu2023visual}. 
Our model is trained entirely on synthetic trajectories in Overcooked, to facilitate diverse behavior exploration. 
We collected a set of gameplay trajectories by deploying synthetic users in procedurally generated overcooked configurations.
We then utilized these trajectories to curate two types of datasets; (1) Grounding datasets focused on enabling the model to accurately interpret the observation space and (2) Task-specific datasets focused on analyzing and correcting defective user-behavior. 
These data span a wide range of behavior types and task configurations, enabling clear stratification and a direct evaluation of the assistive model’s generalization capabilities.

Our evaluations seek to analyze the generalization capabilities of an assistive foundation model along multiple axes. 
First, we evaluate on withheld defective categories and find that, in both zero-shot and few-shot settings, our trained model outperforms a GPT-4o–based behavior critic at predicting corrective feedback and actions. 
Next, we study task-level generalization, where the model must provide open-ended assistance on entirely new recipes. This task is extremely difficult, as it requires the model to (1) compose learned individual concepts in completely new ways, such as learning to fill a soup pot with steaks instead of tomatoes, and (2) coach a user on how to perform this completely new task more efficiently. In this demanding setting, our approach outperforms the GPT-4o baseline when providing corrective actions in the zero-shot setting and achieves superior performance on both feedback and corrective action prediction tasks when exposed to few-shot training examples.
% Together, these results demonstrate that carefully constructed assistive data can enable robust generalization to novel tasks, behaviors, and deployment settings. 
% Crucially, we include an investigation into when such approaches break down, highlighting the limits of data for open-set interactive assistance tasks.
Together, our results show that carefully constructed assistive data can enable robust generalization, while also revealing clear limits of data-driven approaches for open-set interactive assistance.
Our contributions can be summarized as follows:
\begin{enumerate}
    \item We develop a synthetic data generation framework that simulates long-horizon user trajectories in Overcooked, producing complementary datasets that endow an embodied foundation model with grounding, actuation, and assistive capabilities. 
    \item We train a multimodal model to perform Open-Set Assistance, and show that the model, leveraging diverse assistive data along with an LLM backbone can generalize to unseen task configurations and defective categories. 
    % \item We detail a framework for training such an assistive agent, including a synthetic data generation pipeline and training curriculum to elicit general purpose assistance for low-level actions and high-level knowledge gaps. 
    \item We derive insights into effective dataset design for embodied assistance, highlighting the role of multimodal compositionality, spatial reasoning, and task decomposition.
    % \item We compare our model to an off-the-shelf behavior critic (GPT-4o), which can similarly perform open-set assistance, and outperform this baseline on both on novel defective characteristics and task-configurations. 
\end{enumerate}

\section{Related Work}
\label{sec:related_work}

\subsection{Data Generation for Embodied Foundation Models}
Synthetic data are used primarily to fill in gaps in underrepresented regions of the data distribution, such as missing domains~\cite{chen2025robotwin20scalabledata} or safety-critical long-tail settings~\cite{ding2023}.
The proficiency of SOTA language and visual models has allowed researchers to adapt insights from these models to scale up the grounding, multimodal understanding, and actuation capabilities of ``in-the-wild'' embodied foundation models~\cite{ahn2024autort, Tang2025, mandlekar2023mimicgen, singh2024synthetica}. 
Another approach is to directly optimize the data generation process by indexing key performance metrics in a downstream task~\cite{huang2025subjectdrive, wang2024robogen}.
In addition to modeling based approaches, recent research has  proposed numerous novel simulation mechanisms to allow accelerated data generation for tasks that were once considered infeasible~\cite{DEMELO2022174, li2024egogen, chen2025, yang2025drivearena}.
These methods highlight the significant utility of targeted data generation in systematically addressing deficiencies in model capabilities and improving generalization across tasks. 
% synthetic data in scaling the capabilities of embodied foundation models. 
% In this paper, we seek to study a

\subsection{Simulating user-behavior in Sequential Decision Making Settings}
Modeling user-behavior is a critical component of an assistive model. 
The first component of user-modeling is simulating user-behavior in an embodied environment. 
A popular method of building synthetic agents has been through co-play or behavior cloning~\cite{strouse2021collaborating, yan2023efficient, carroll2019utility, wang2023utility}.
Other methods seek to directly model a distribution of user-actions which take into account style and preferences, given an input-state~\cite{pearce2023imitating, laidlaw2022boltzmann}.
Finally, a more recent suite of methods utilize LLMs to function as user-models which condense user-preferences through demonstrations or added-context~\cite{yun2025sleepless, zhang2023large, zhang2024proagent}. 
These methods provide critical tools for simulating interactions with complex synthetic user's to build models capable of understanding and adapting to human-behavior~\cite{anthis2025llm}. 
% user simulation for co-play - ~\cite{strouse2021collaborating, yan2023efficient}
% user simulation via learning a human policy - ~\cite{pearce2023imitating, laidlaw2022boltzmann, carroll2019utility}
% llm as user model - ~\cite{zhang2023large, zhang2024proagent, anthis2025llm, Zhang_Liu_Liu_Zhong_Cai_Zhao_Zhang_Liu_Jiang_2025}
% synthetic users - ~\cite{yun2025sleepless}
% cognitive modeling - ~\cite{}
\subsection{Assistive Modeling}
Addressing user-deficiencies in embodied settings requires inferring the user's high-level strategy, and intent~\cite{hoffman2024inferring}, identifying any deficiencies, and generating appropriate feedback. 
Relevant prior work includes methods to discover user-traits from their underlying policies~\cite{ankile2023discovering}, or condition the agent's policy on an estimated distribution over the partner's intention~\cite{decastro2024dreaming, wang2023utility, zhao2022coordination}. 
Recent work has also developed bespoke Theory-of-Mind modules which provide cognitive scaffolds to ground predictions of human-behavior~\cite{liu2024heirarchical, cross2024hypothetical}. 
Upon identifying a user's intention, a model can then move on to correcting the deficiencies in their behavior; either through language-based critique or corrective guidance. 
Behavior critique methods process behavior trajectories and generate actionable feedback or reflections~\cite{liu2023reflect, guan2024task, gopinath2024computational}. 
Behavior corrections provide guidance actions by via a behavior model, or select a corrective strategy, to physically demonstrate correct behavior~\cite{verghese2025user, yang2024trajectory, sinha2024real}.

Despite substantial progress in synthetic data generation, user simulation, and assistive modeling, existing work often studies these components in isolation and relies on closed sets of corrective strategies, external planners, or real-world data. Furthermore, testing an assistive model's ability to generalize corrective guidance across diverse unseen tasks and failure modes remains largely unexplored~\cite{zapata2025advancements}. In this work, we introduce \textit{open-set assistance} as a challenging testbed for evaluating such generalization and for informing the design of future embodied assistive models.

\section{Open-Set Corrective Assistance}
\label{sec:methodology}
% There are broadly two categorizations of corrective assistance an embodied agent can provide to assist a user in a human-AI interaction, i.e. (1) Language-based coaching to empower the user to understand their deficiencies and how to correct those deficiencies~\cite{liu2023reflect, guan2024task}, or (2) Physical actuation correct user actions by demonstrating a remedial action to guide the user in the right direction~\cite{verghese2025user, yang2024trajectory}. 
% % to take over from the user to demonstrate a remedial action to guide the user in the right direction. 
% Prior work has primarily relied on approaches which simplify the problem by enabling their model to classify from a fixed set of categories of coaching categories or corrective actions. 
In this work, we seek to build a model which can provide Open-Set Corrective Assistance, i.e. our assistive model needs to predict either coaching-based assistance or correction-based guidance based on the instruction provided, and the model is never given a preset list of possible actions or feedback classes. 
In the rest of this section, we define the task of Open-Set Corrective Assistance($\S$~\ref{sec:task-def}), and provide details regarding our model architecture($\S$~\ref{sec:model-arch}) to facilitate such assistance. 

\subsection{Problem Definition}
\label{sec:task-def}
In this paper, we seek to build an assistive model capable of inspecting multimodal traces of user-behavior, and generatively providing assistance, either through language or physical actuation. 
% To perform this task, we fine-tune a pretrained Llama-3 model, $M$, and a ViT encoder, $E$, to provide assistance, $a^{l|c}$. 
We train our model on a dataset of multimodal trajectories in Overcooked, along with the assistance in the form of language or a physical action, i.e. $\mathcal{D}_{train} = \{\tau_i, [c|a]\}_{i=1}^{|\mathcal{D}_{train}|}$. 
Each trajectory, $\tau_i$, is comprised of an interleaved sequence of state-action pairs ($<s_j^i, a_j^i>$), where the states are RGB images of the overcooked game-state, and actions are represented by language, i.e. ``move left,'' ``no action,'' ``interact,'' etc.
$[c|a]$ represent the modalities of assistance the model can provide which is either language-based coaching advice ($c$), or a remedial action ($a$). Note that coaching advice ($c$) is delivered in a terminal setting, whereas remedial actions ($a$) are generated only after the full input trajectory has been observed, rather than in an intermittent manner. 

Each trajectory in $\mathcal{D}_{train}$ is a user's gameplay history with an underlying defective category, $\delta_k \in \Delta$, where $\Delta$ corresponds to the set of all defective categories. 
Defective categories, or \textit{defects}, are defined as an impairment in or characteristic of a user's behavior which may adversely impact cognitive processes such as attention, memory, decision-making, planning, etc. 
In this paper, we focus on defects that fall under the category of  ``cognitive impairment'', in that they are short-term conditions that impair decision-making, rather than a condition that is progressive and permanent~\cite{dhakal2020cognitive}. 
Specifically, we focus on defects that manifest as improper planning/sequencing of operations, or impairments in visuo-spatial skills leading to difficulty in identifying spatial relationships and interacting with the environment. 
Examples of some defects that we design are ``Player does not know when the tomatoes in the pot have started cooking,'' (Visuo-spatial) or ``Player thinks it is best to only serve a single dish'' (Planning). 
Our training dataset has a total of seventeen defects, $|\Delta| = 17$, where one of the possible defective categories is a null setting, i.e., "No Defect." 
% Despite the limited number of defect categories, the manifestations of these defects in actual gameplay trajectories can be extremely diverse because our data generation pipeline features randomly generated layouts and objectives (\ref{sec:syn_data_gen}). 
Although we do not work with an infinitely large set of defective categories, we selected a wide range of defects that encompass the variety of possible behaviors exhibited by an overcooked player. 
The full set of defects is included in the Appendix ($\S$~\ref{sec:defects}).

We aim to train a unified model comprised of a base LLM, $\pi_\theta$, and a visual encoder, $\phi$, to provide corrective guidance, i.e. $c_i \sim \pi_\theta(\cdot | \phi(\tau_i), p_t)$. 
$p_t$ represents the associated task prompt for either action-prediction or feedback-prediction. For the action-prediction task, $p_t$ also includes a reference trajectory of the user's gameplay on a simple map, $\tau_r$ to clarify the user's general strategy (described in further detail in $\S$~\ref{sec:syn_data_gen}) and allow the model to produce guidance aligned with the user's overarching strategy. 
The full model-instructions are provided in the Appendix, $\S$~\ref{sec:model-instructions}). 

\subsection{Model Architecture}
\label{sec:model-arch}
We employ a projection-based multimodal model-structure akin to Llava~\cite{liu2023visual}. 
We utilize Llama-3~\cite{grattafiori2024llama} as our base-LLM, and a ViT-base model~\cite{dosovitskiy2020image} as our image-encoder. 
The image-encoder generates visual features for the entire batch of images in the trajectory, $Z_i = \phi(s_i)$.
We then utilize an language-projection layer, $W_p$, to project the image embeddings into the space of the language model, $X^s_i = W_p\cdot Z_i$. 
These image-tokens, $X^s$ are interleaved with the language embeddings from the LLM, $X^a$ to make up the entire set of trajectory tokens which are input into the LLama model, $X^\tau = X^s \otimes X^a$.
For decoding, we employ a text-only decoding framework where both the coaching and corrective-actions are decoded as language tokens. 

\section{Data Definitions}
\label{sec:data_gen}
We synthetically generate a dataset of trajectories to train our model to perform Open-Set corrective assistance.
% Our training data for open-set assistance is comprised of defective trajectories with appropriate corrective guidance (as per $\S$~\ref{sec:task-def}). 
Our goal is to derive clearly stratified synthetic data to accurately test our model's ability to adapt to new tasks and defective behaviors. \color{black}
We begin by describing the strategies employed to generate each component of our dataset, i.e. trajectories, corrections, etc. ($\S$~\ref{sec:syn_data_gen}).
We then conclude with a comprehensive description of each of our training sets used during model-training ($\S$~\ref{sec:datasets})

\subsection{Data Generation}
\label{sec:syn_data_gen}
\textbf{Synthetic Users.} Our goal was to generate a diverse dataset which covered a range of varying strategies for playing Overcooked. 
We first designed an ``Overcooked API'' which provided skills to our synthetic agents e.g. $\texttt{pickup(dish, counter)}$, $\texttt{move\_to(sink)}$, etc. 
Using this API, we created a set of 5 rule-based heuristics ($\mathbb{H} = \{H_1, H_2, H_3, H_4, H_5\}$), where each heuristic represents a different set of priorities, or strategies for employing a task, e.g. \textit{``Always prioritize steak preparation over soup preparation''}, or \textit{``Always make sure dishes are cleaned and ready before starting any meal-prep''}, etc. 
Note that each of these heuristics were designed to represent diverse preferences for playing  Overcooked, they were not designed to be optimal. 

\textbf{Defective Rollouts.} We instrumented our agent APIs with defective wrappers to enable altering of a specific aspect of their behavior, such as ``the user no longer being able to identify when a cooking pot is full,'' or ``the user ignoring trip hazards during their path planning.''
We roll out trajectories, $\mathcal{T}$, with these defective wrappers and heuristics, injecting additional stochasticity so the synthetic user takes a random action with 20\% probability.
% Each synthetic user and defect pair is run once on each overcooked map. 
To ensure diversity of rollouts, we apply noisy rollouts, wherein there is a small probability of a random action being selected in the trajectory at every step. 
This noise pushes the player into atypical states, yielding more varied behaviors across trajectories. 
Furthermore, to prevent overfitting to repetitive patterns, each user–defect pair is executed only once in each map configuration.
Our training dataset uses a set of 450 procedurally generated maps during the rollouts (The prompt utilized to generate the map-configurations is provided in the Appendix, $\S$~\ref{sec:map-prompt}).

\textbf{Ground Truth Correction Generation.} We seek to assign a ground truth corrective-action or coaching snippet to each trajectory in the dataset. 
The corrective action is computed by predicting the next action of the heuristic, given the last state, without the defective wrapper. 
The ground truth coaching snippets are generated synthetically using an independent GPT-4o model, $M$. 
% Our synthetic pipeline to generate coaching follows three steps. 
First, we create a set of boilerplate coaching snippets, $C^{raw} = \{M(\delta) ~~  \forall \delta \in \Delta\}^N$, by prompting the language model, $M$, to generate a coaching snippet which identifies and corrects a given defect, for each defect in the dataset. 
This process is repeated $N$ times per defect, which in our case is set to 5. 
Next, for each seed coaching snippet, we use $M$ to tailor the response to a specific ``persona'' ($p \in \mathcal{P}$), to expose the model to a more diverse set of coaching snippets.  
% Next, we prompt the language model to adopt ``personas'' to encourage the model to produce more diverse and varies forms of coaching. 
Each persona adapts the raw coaching snippet to exhibit a specific style such as ``urgency,'' ``frustration,'' ``encouragement,'' ``brevity,'' etc., $C_\delta^{final} = \{ M(p,c^{raw}) ~~ \forall p \in \mathcal{P} ~\text{and} ~ \forall c^{raw} \in C_\delta^{raw} \}^N \forall \delta \in \Delta$. 
In between each of these steps, we adopt a self-evaluation step wherein the model validates whether the generated coaching snippet  adequately addresses the original defect. 
This validation is done in an ensemble fashion, where the same language model is asked to vote five times, a data-point is only selected if at least four out of five instances passed the validation test. 
Finally, each trajectory in our corpus is randomly assigned one of the synthetically generated coaching snippets corresponding to the defect in the given trajectory. 

\textbf{Reasoning Trace Generation}
Finally, we augment the trajectories in our dataset with synthetic reasoning traces.
The inclusion of reasoning traces in our dataset enables us to measure whether teaching our model to reason on the behavior exhibited in a trajectory can improve it's ability to generalize to new defective categories or assist on previously unseen tasks. 
We collect reasoning traces by prompting $M$ to analyze a trajectory by parsing an abstract list of events that occurred in the trajectory. 
These abstracted events, $E(\tau_i)$, include any important outcomes or changes in the state, based on the player's actions and are computed during the initial trajectory rollout. 
% The abstracted events during a trajectory, , are computed during the initial trajectory rollout.  
Each reasoning trace includes three components: (1) A \textit{summary} of the trajectory describing what the user tried to do, (2) A description of the player's \textit{successes} over the course of the trajectory, and (3) The set of \textit{challenges} faced by the player while performing their task. 
The language model performs this reasoning procedure, $r_i^{sum}, r_i^{suc}, r_i^{ch} \sim M_{reason}(E(\tau_i))$,  three times for each example, and chooses the best output for each category based on a self-assessed measure of confidence with regards to the output's relevance to the original set of events. 

\subsection{Training Datasets}
\label{sec:datasets}
We curate two sets of datasets for model-training: (1) Grounding Datasets - Datasets which equip the model with the ability to ingest a multi-modal trajectory and ground the events which occur in the trajectory, and (2) Task-Specific Datasets - Datasets which teach the model to analyze trajectories, identify defects and provide assistance. 

\subsubsection{Grounding Datasets}
We generate visual question-answering (VQA) datasets, by sampling images or sequences of images from trajectories generated by our synthetic agents. 
We designed questions to enable the model to visually analyze the observation space to make assessments about a given state or identify important events which occurred across a sequence of states. 
We design three VQA datasets for this task, 
\begin{itemize}
    \item Image-QA: In the Image-QA dataset, $\mathcal{D}^I = \{I_i, \{q_j, a_j]\}_{j=1}^6\}_{i=1}^N$, each datapoint is comprised of an image, randomly sampled from a trajectory of a synthetic heuristic, associated with six question-answer pairs associated with the given state, which are randomly sampled from a larger set of questions. These questions probe the model's grounding of spatial information  in the environment, which is a task MLLMs often struggle with~\cite{rajabi2024gsr, zhang2025mllms}.  The entire dataset is comprised of 55,000 examples. 
    \item Trajectory-QA: The Trajectory-QA dataset is a dataset comprised of trajectories, with six associated question-answer pairs, $\mathcal{D}^T = \{ \hat{\tau}_i, \{q_j,a_j\}_{j=1}^6 \}_{i=1}^N$.
    The trajectory in each datapoint, $\hat{\tau}_i$, is selected by extracting a sub-trajectory from a full-rollout, of length, $K \sim \mathcal{N}(15, 2)$. 
    These questions seek to improve the model's temporal grounding capabilities, i.e., identify how the actions taken in the trajectory affect the environment. This dataset has a total of 54,000 datapoints. 
    \item Video-QA: Similar to the previous dataset, the Video-QA dataset, $\mathcal{D}^V$ is comprised of trajectories augmented with question-answer pairs. However, in this dataset, the trajectories are just a sequence of images in the form of a video, $\hat{\tau}_i = \{s_k\}_{k=1}^K$. We utilize the same template for the question-answer pairs as that of the Trajectory-QA dataset. The purpose of the dataset is to ensure that the model adequately relies on its visual encoding capabilities rather than overfitting to the sequence of actions provided in the trajectory. This dataset is comprised of 55,000 datapoints. 
    % The length of each trajectory, is randomly selected for each sample, by sampling from a normal distribution with a mean of fifteen and a standard deviation of two. 
\end{itemize}
The entire set of questions for each of these grounding datasets can be found in Appendix~\ref{sec:vqa-questions}. We label the set of all these gounding datasets as $\mathcal{D}_{ground}$.

\subsubsection{Task-Specific Datasets}
By utilizing our synthetic data generation procedures described in $\S\ref{sec:syn_data_gen}$, we assemble three datasets which directly pertain to corrective assistance (see Appendix Figure~\ref{fig:dataset_diagram}). 
\begin{itemize}
    \item \textbf{Coaching}: We formulate the coaching task using the dataset, $\mathcal{D}^{\text{coach}} = \{ \tau_i^{1:t}, \; c_{\delta_i}, \; [r_i] \}$
% \[
% \mathcal{D}^{\text{coach}} = \{ \tau_i^{1:n}, \; c_{\delta_i}, \; [r_i] \}, 
% \quad c_{\delta_i} \sim C^{\text{final}}_{\delta_i}
% \]
where each datapoint consists of a trajectory $\tau_i^{1:T} \in \mathcal{T}$, 
a coaching signal $c_{\delta_i}$, which is randomly selected from the set of available coaching snippets for the given defect ($C^{\text{final}}_{\delta_i}$), and an optional reasoning trace $[r_i]$ associated with the trajectory. 
Note that the coaching snippet implicitly aligns with the behavior exhibited in the trajectory, since the trajectory is generated by applying a particular defect, but this alignment is not explicitly enforced while assigning the coaching snippet to the trajectory. This set contains 26,000 training examples.
\item \textbf{Corrections}: We formulate the correction task as follows, $\mathcal{D}^{\text{correct}} = \{ \tau_{ref}, \tau_i^{t-10:t}, \; \hat{a}_{t+1}, \; [r_i] \}$.
% \[
% \mathcal{D}^{\text{correct}} = \{ \tau_{ref}, \tau_i^{n-10:n}, \; \hat{a}_{n+1}, \; [r_i] \}
% \]
The model needs to predict the next \textit{corrected} action the user should take to steer them away from their defective gameplay. 
% The key difference between the coaching and corrections task pertains to the inclusion of an additional reference trajectory. 
For the corrections task, the model is also provided a reference trajectory,  $\tau_{ref}$, comprised of a rollout of the specific user acting in a small map (different to the any of the maps utilized to simulate trajectories) to highlight the user's high-level strategy.
% To aid the model in providing assistance catered to the user's specific gameplay style, they are provided with a reference trajectory, $\tau_{ref}$, comprised of a rollout of the specific user acting in a small map (different to the any of the maps utilized to simulate trajectories) to highlight their high-level strategy. 
Finally, we restrict the input trajectory to the 10 preceding steps, which allows the reference trajectory to fit within the model's available context window. This dataset contains 27,000 training examples.
\item \textbf{Defect-Delineation}: Our third task task, Defect Delineation, seeks to improve the model's ability to discern how different defects manifest in user behavior. 
This task was inspired by the Next-Sentence Prediction pretraining task~\cite{devlin2019bert}, leveraged to build a greater semantic understanding of sentence-similarity. 
Similarly, in Defect-Delineation the model inspects two trajectories and predicts whether  the two trajectories have the same defect, and also the specific defect in each trajectory, 
$\mathcal{D}^{DD} = 
    \big\{ \tau^{(1)}, \; \tau^{(2)}, \; \delta^{(1)}, \; \delta^{(2)}, \; 
    \mathbb{I}\{\delta^{(1)} = \delta^{(2)}\} \big\}_{i=1}^{|\mathcal{D}^{DD}|}$. 
Each trajectory is randomly sampled from our dataset of trajectories, $\mathcal{T}$.
The output is formatted as a text-based answer, ``Yes. The defect in both trajectories is <defect>'' or ``No. The defect in trajectory 1 is <defect1>. The defect in trajectory 2 is <defect2>.''
Our dataset includes five negative samples for every positive sample, as the task of distinguishing and correctly predicting the defects in two different trajectories is more challenging. This dataset contains 20,000 training examples.

\end{itemize}
Our default training dataset, $\mathcal{D}_{train}$, includes all three of these datasets with equal weighting. 

\section{Experiments}

\subsection{Baselines}
% We compare our synthetically trained model against a state-of-the-art baseline designed to exhibit strong generalization capabilities.
A suitable baseline for Open-Set Corrective Assistance must be able to process video trajectories, perform visual grounding to interpret task instructions, and provide the corresponding assistance.
We adopt a methodology for behavior correction, developed in prior work, which empowers pretrained VLMs to provide language-based coaching based on videos of a robot's policy~\cite{guan2024task}.
% In prior work, the authors leveraged GPT-4o to generate language-based coaching. 
In our work, we extended this baseline to generate corrections as well as coaching. 
\begin{enumerate}
    \item Behavior Critic (GPT-4o) - We prompted GPT-4o to perform assistance based on a downsampled trajectory. In addition to the trajectory, the model is provided the rules of overcooked for additional grounding. Furthermore, unlike our model, which needs to implicitly identify new defects without any prior knowledge, the GPT-4o behavior-critic is also provided the full list of potential defective behaviors that could be found in the behavior. 
    \item Behavior Critic (GPT-4o) + language summary - We extend the original behavior critic by adding reasoning traces from our dataset to the prompt as a part of the input to help improve the model's ability to identify the defect and provide assistance. 
    % This baseline allows us to better understand two concepts; (1) Are the synthetic reasoning traces in our dataset actually helpful for assistance? and (2) Can our model outperform this version of GPT-4o with additional grounding information about the trajectory? 

\end{enumerate}

The prompts for both of these baselines are included in the appendix($\S$~\ref{sec:behavior-critic}).

\subsection{Evaluation Tasks}
For the novel task-formulation of open-set assistance, our evaluations seek to comprehensively examine how well our assistive model can perform across multiple axes of generalization. 
In all our experiments (unless otherwise stated), we perform few-shot training to fine-tune our model on a small set of examples prior to testing on held-out data (10 examples per new defect). 
\begin{itemize}
    \item \textbf{Novel Defective Modalities} - This evaluation tests how well our model can utilize its understanding of the task to  provide actionable assistance in previously unseen failure modes (defects). 
    % is able to adapt to defective manifestations that the model did not see during training. A successful assistive model will be able to utilize its understanding of the task to categorize the new failure modes and provide actionable assistance. 
    For instance, during training, the model may encounter a defective modality such as “Player believes it is best to serve only soup,” while a held-out test defect might be “Player believes it is best to serve only steak.”
    If the model has learned the underlying concept of a player omitting key components of a recipe, it should be able to generalize this understanding to novel contexts involving different missing components. 
    \item \textbf{New Tasks (Overcooked Recipes)} - During training the agent is always following a single recipe. In this evaluation, we measure how well the model is able to assist users performing tasks, i.e. completing two new recipes, within Overcooked. This evaluation probes the model’s ability to adapt its domain knowledge to a related task within the same domain through compositionality, grounding unseen recipes to new object configurations and layouts to identify previously unseen failure modes.
    All recipes are provided in Appendix~\ref{sec:recipes}.
. \color{black} 
\end{itemize}

Finally, we perform two ablations pertaining to the training set configuration. First, we compare multi-dataset training on the full $\mathcal{D}_{train}$ against single-task training on $\mathcal{D}^{coach}$ or $\mathcal{D}^{correct}$ to assess whether jointly learning related assistive tasks leads to more robust and adaptable assistive behavior. 
Second, we assess whether co-training with our grounding datasets ($\mathcal{D}_{ground}$) improves the model’s ability to localize actions and events required for novel recipes and translate them into assistance. \color{black}

% \subsection{Research Questions}
% \begin{enumerate}
%     \item Can our model learn to effectively assist in settings with novel defective categories?
%     \item Can our model learn adapt to new tasks within the environment?
%     \item How does reasoning improve our assistive model's zero-shot adaptation to 
%     \item How does multi-task training contribute to the model's ability to generalize to new tasks/defects? 
% \end{enumerate}

\subsection{Metrics}
We utilize two metrics to measure the accuracy of the model for each modality of assistance. 
\begin{itemize}
    \item LLM-as-judge (Coaching) - We leverage an LLM to identify whether or not a generated coaching snippet adequately corrects the underlying, ground-truth defect. The full prompt used for enabling GPT-4o to evaluate our model-outputs is included in the appendix. 
    \item Accuracy (Corrections) - To evaluate the correction-generation capabilities of our model, we simply measure the accuracy when compared to the ground truth action. 
\end{itemize}

\subsection{Results}

\begin{table}[t!]
\centering
\caption{
\textbf{Generalization across defects and tasks.}
We compare assistive performance when generalizing to (left) unseen defective behaviors and (right) novel task configurations (recipes). 
Models trained entirely on diverse assistive data outperform behavior-critic baselines when adapting to new defects with only a few examples, with performance saturating at the 1B scale. 
In contrast, generalization to new tasks benefits substantially from model scaling, likely due to the stronger multi-modal grounding capacity offered by the larger model.}
\label{tab:main}
\begin{tabular}{lcccc}
\toprule
& \multicolumn{2}{c}{\textbf{Held-Out Defects}} & \multicolumn{2}{c}{\textbf{Task Generalization}} \\
\cmidrule(lr){2-3} \cmidrule(lr){4-5}
\textbf{Baseline} & \textbf{Coaching} & \textbf{Corrections} & \textbf{Coaching} & \textbf{Corrections} \\
\midrule
Behavior Critic & 21.00 & 20.40 & 34.21 & 9.17 \\
Behavior Critic + Summaries & 55.70 & 19.80 & 71.05 & 15.83 \\
Ours 1B & 76.60 & \textbf{55.70} & 50.88 & 50.83 \\
Ours 8B & \textbf{77.80} & 54.60 & \textbf{85.96} & \textbf{56.67} \\
\bottomrule
\end{tabular}
\end{table}

\begin{table}[t!]
\centering
\caption{The 8B variant of our model outperforms the behavior critic baseline in all settings except while coaching in the task generalization setting. Coaching users to perform new tasks proves to be a challenging setting for the model, however, the inclusion of reasoning traces provides an 8\% boost in performance. }
\label{tab:zero-shot-combined}
\begin{tabular}{lcccc}
\toprule
& \multicolumn{2}{c}{\textbf{Held-out Defects}} & \multicolumn{2}{c}{\textbf{Task Generalization}} \\
\cmidrule(lr){2-3} \cmidrule(lr){4-5}
\textbf{Baseline} & \textbf{Coaching} & \textbf{Corrections} & \textbf{Coaching} & \textbf{Corrections} \\
\midrule
Behavior Critic & 21.00 & 20.40 & \textbf{34.21} & 9.17 \\
Ours 8B & \textbf{34.30} & \textbf{39.70} & 0.00 & \textbf{55.00} \\
Ours 8B + reasoning & 18.10 & 32.60 & 8.77 & 43.30 \\
\bottomrule
\end{tabular}
\end{table}

\textbf{Our model can learn to assist users with unseen defective behaviors with just 10 training examples per defect} -- 
Table~\ref{tab:main} reports the performance of our model on the dataset comprised of trajectories with novel defective behaviors. 
First, we find that our model, when trained on the three task-specific datasets ($\mathcal{D}^{coach}, \mathcal{D}^{correct}, \mathcal{D}^{DD}$) outperforms an off-the-shelf GPT-4o model for both tasks. 
Incorporating the visual-reasoning traces as an input to GPT-4o's input, boosts the model's ability to accurately coach by $34\%$, however, it is still inferior to our model. 
Upgrading the base LLaMA model from one billion (1B) to eight billion (8B) parameters results in only negligible performance differences (+1 for coaching and –1 for corrections). 
This suggests that the model has reached a saturation point on the dataset for the held-out defect evaluation.

\textbf{Our model can generalize to new in-domain tasks} -- 
We report findings on our evaluation of the model's ability to assist users performing new recipes in Table~\ref{tab:main}. 
First, our 1B model surpasses the base behavior critic (+15/+41\%) but falls short of the summary-augmented version (-20/+35\%). 
Unlike in the case of the held-out defect evaluation, performance of the model increases, for both corrections and coaching, when scaling up from the 1B LLaMa model to the 8B LLaMa model. 
This indicates that a greater degree of multimodal compositionality is required for this task, which larger VLMs are better equipped to handle.  

\begin{table}[h!]
\centering
\caption{
\textbf{Effect of dataset ablations on assistive generalization.}
% We compare the performance of collective training on $\mathcal{D}_{train}$ to individually training on $\mathcal{D}^{coach}$ or $\mathcal{D}^{correct}$. 
Training jointly across coaching, correction  and defect delineation improves downstream assistive performance compared to individually training on $\mathcal{D}^{coach}$ or $\mathcal{D}^{correct}$.
}
\label{tab:ablation-combined}
\begin{tabular}{lcccc}
\toprule
& \multicolumn{2}{c}{\textbf{Unseen Defects}} & \multicolumn{2}{c}{\textbf{Novel Tasks}} \\
\cmidrule(lr){2-3} \cmidrule(lr){4-5}
\textbf{Training Data} & \textbf{Coaching} & \textbf{Corrections} & \textbf{Coaching} & \textbf{Corrections} \\
\midrule
1B ($\mathcal{D}_{train}$) & \textbf{76.60} & \textbf{55.70} & 50.88 & \textbf{50.83} \\
1B ($\mathcal{D}^{coach}$) & 38.10 & -- & \textbf{54.39} & -- \\
1B ($\mathcal{D}^{correct}$) & -- & 29.40 & -- & 33.34 \\
\bottomrule
\end{tabular}
\end{table}

\textbf{Inducing reasoning priors for assistance improves zero-shot coaching performance while generalizing to new tasks } --
% Table~\ref{tab:zero-shot-combined} reports the zero-shot performance of our assistive model (at an 8B scale) with and without reasoning. 
% First, we provide evidence that the reasoning traces used to train our model are beneficial towards improving assistive performance. 
% The addition of the summary segments to the behavior critic generally increases asssistive performance on both tasks, however the gains are more substantial for the coaching task than the corrections task. 
When our model is evaluated in a zero-shot setting, we see a reduction in performance on the coaching task for both sets of evaluations (see Table~\ref{tab:zero-shot-combined}). 
Our zero-shot model still outperforms the behavior critic on both assistive tasks on the held-out defect evaluation. 
However, when reasoning is added, model performance worsens. 
This is likely due to mode-collapse; since the model has not been trained to reason with reinforcement learning, the model may struggle to condition its assistance on out-of-distribution reasoning traces, even if the reasoning traces correctly ground the input trajectory. 
For the task-generalization evaluation, the model outperforms both versions of GPT-4o while predicting corrections, however, completely fails to coaching while assisting on novel-tasks ($0.00\%$). 
Coaching the user to perform new tasks is much more challenging for any assistive model, as the model needs to infer mistakes from the new recipes and map failures identified in these trajectories to these inferences. 
Without any training on held-out data, the model is unable to make these inferences. Notably, incorporating reasoning traces mitigates this issue, improving coaching performance by $8\%$, however, the model is still unable to match the behavior critic's performance on task-generalization.  
% \color{blue}
% We provide some example outputs and reasoning traces of our model in the Appendix.
% \color{black}

\textbf{Training on multiple datasets simultaneously improves the model's ability to assist} --
Table~\ref{tab:ablation-combined} presents a dataset-analysis to understand the effect of each dataset on the model's ability to perform open-set assistance. 
These experiments are performed with our 1B-scale model. 
First, we measure the impact of multi-task training with all three datasets in $\mathcal{D}_{train}$. 
Having the model simultaneously train on both tasks, as well as the auxiliary defect delineation task, wherein the model is prompted to identify and delineate defects, generally improves downstream performance on both tasks. 

\begin{table}[t!]
\centering
\caption{\textbf{Inclusion of additional grounding datasets improves model's ability to assist in new task configurations.} This figure depicts the performance of the assistive model, on the task-generalization experiment, after co-training with varying configurations of the grounding datasets, $\mathcal{D}_{ground}$. Among the grounding datasets, the Trajectory-QA dataset, $\mathcal{D}^T$, appears to be the most effective, whereas the Video-QA dataset, $\mathcal{D}^V$, appears to reduce performance.}
\label{tab:ablation-task-gen}
\begin{tabular}{lcc}
\toprule
\textbf{Baseline} & \textbf{Coaching} & \textbf{Corrections} \\
\midrule
1B  & 50.88 & 50.83 \\
1B ($\mathcal{D}_{ground} + \mathcal{D}_{train}$) & 58.77 & \textbf{53.33} \\
1B ($\mathcal{D}^I + \mathcal{D}_{train}$)  &  69.30 & 46.67 \\
1B ($\mathcal{D}^V + \mathcal{D}_{train}$)  &  46.49 & 42.50 \\
1B ($\mathcal{D}^T + \mathcal{D}_{train}$)  &  \textbf{78.95} & 52.50 \\
\bottomrule
\end{tabular}
\end{table}

\textbf{Co-training with grounding datasets can improve visual compositionality, thereby benefiting task generalization - }
We hypothesized that task generalization requires stronger visual compositionality to translate unseen behaviors into effective assistance. To test this, we co-train with the grounding datasets, introduced in Section~
\ref{sec:datasets}, and find that such co-training, both jointly and individually, improves downstream assistive performance on novel tasks (Table~\ref{tab:ablation-task-gen}).
% This supplements our findings related to reasoning while assisting user's to perform new recipes. 
Improved performance may arise from the model’s increased ability to recognize key in-game events, such as the completion of a delivery or proximity to the delivery location, which in turn supports the inference of new failure modes from trajectories of players performing unfamiliar tasks. 
Among the grounding datasets, $\mathcal{D}^T$ appears to be the most effective, whereas co-training with only $\mathcal{D}^V$ seems to reduce task performance. 
This finding implies that access to action sequences may be critical for analyzing new failure modes in novel recipes.
$\mathcal{D}^I$ also appears to provide benefits; unlike the Yes/No questions in the Trajectory and Video datasets, the Image-QA dataset requires more complex analyses of the game state, including path planning and object disambiguation, which may improve the model’s ability to process image-based states.

\section{Discussion}
This work examines the capabilities of diverse assistive data in training foundation models to perform  embodied open-set assistive tasks. We generate a synthetic dataset and train a model to perform a representative task, Open-Set Corrective Assistance, in which the model must provide corrective guidance by observing embodied trajectories of user behavior, without access to a closed set of corrective categories. We evaluate models, trained entirely on synthetic data, in two complementary settings: (1) generalization to trajectories containing defective behaviors not seen during training, and (2) generalization to novel task configurations (recipes) in Overcooked. 

Firstly, our dataset ablations yield several insights into the composition of  training data required to support generalization. In particular, task-level generalization places greater demands on multimodal compositionality, requiring  data that captures higher-level relationships between actions, goals, and environmental outcomes. We find that multi-task co-training across coaching, correction, and defect delineation consistently improves downstream assistive performance, while grounding datasets that contain re-usable information pertaining to action outcomes are especially beneficial for generalization to connect outcomes to deficiencies in new task configurations. Additionally, inducing reasoning priors through behavior summaries improves assistive performance, particularly for coaching and zero-shot evaluation. However, performance degrades when reasoning traces fall outside the training distribution, indicating that current forms of reasoning supervision are brittle. Together, these results suggest that effective training datasets for assistive modeling should emphasize decompositional structure, explicitly separating perception, interpretation, and actuation,  rather than relying solely on end-to-end demonstrations.

\section{Limitations}
Finally, there are two important limitations to cover pertaining to pending avenues of exploration and model-capabilities. 
First, it is important to highlight that the capabilities of a diverse trained model to provide assistance on real-human trajectories remains unexplored. 
Having explored the data configurations that is helpful in promoting generalization to novel tasks and defective behaviors, a critical next step is to collect human trajectories to understand the quality, quantity and design of synthetic data required to promote sim2real generalization on complex, embodied assistive tasks requiring foundation models. 
Secondly, our model training procedure leverages a standard instruction-tuning setup, absent of any feedback-based post-training. 
Applying alignment-based approaches could produce assistance that better captures the intrinsic reward model of the user that is being assisted. 
Recently, a suite of methods have been developed to apply reinforcement-learning-based optimization methods to learning in complex multi-turn or embodied settings~\cite{silvastability, ding2025empowering, boyle2025robotxr1}. 
Applying these alignment methods would prove a natural fit to producing assistance that more effectively reasons about user-behavior and thereby exponentiate the grouding capabilities imbued through large-scale training on synthetic data.

\section{Conclusion}
In this paper, we seek to quantify the capabilities of diverse assistance for training assistive model for complex embodied tasks, requiring a multimodal foundation model. 
We study this question through Open-Set Corrective Assistance, a representative and largely unsolved setting in which a model must identify and correct user behavior through language feedback or corrective actions without access to a closed set of defect categories. 
Using Overcooked as our testbed, we collect a set of behavior trajectories by simulating gameplay with a set of synthetic users. 
We then processed these trajectories to produce multi-faceted training data, to imbue the model with grounding, actuation and assistive capabilities. 
Next, we trained our model on these datasets, so as to probe it on two clear stratifications for generalization; (1) Unseen failure modes (defects) and (2) Novel task configurations. 
Through our experiments, we show that our trained model can assist in both these novel environment configurations better than a behavior-critic from prior work. 
Our dataset ablations highlight the importance of designing datasets which individually teach the compounding skills required to perform a task, which in our case were grounding and multi-turn behavior analysis.  
Overall, our work establishes a foundation for future research on open-set assistive models, providing findings beneficial towards extending this work to assistance on real-world interactive data.

\bibliographystyle{unsrt}  
\bibliography{references}  %%% Remove comment to use the external .bib file (using bibtex).

@article{sinha2024real,
  title={Real-time anomaly detection and reactive planning with large language models},
  author={Sinha, Rohan and Elhafsi, Amine and Agia, Christopher and Foutter, Matthew and Schmerling, Edward and Pavone, Marco},
  journal={arXiv preprint arXiv:2407.08735},
  year={2024}
}

@article{sumner2025simcoachcorpus,
  title={SimCoachCorpus: A naturalistic dataset with language and trajectories for embodied teaching},
  author={Sumner, Emily and Gopinath, Deepak E and Dees, Laporsha and Gomez, Patricio Reyes and Cui, Xiongyi and Silva, Andrew and Costa, Jean and Morgan, Allison and Schrum, Mariah and Chen, Tiffany L and others},
  journal={arXiv preprint arXiv:2509.14548},
  year={2025}
}

@inproceedings{li2024egogen,
  title={Egogen: An egocentric synthetic data generator},
  author={Li, Gen and Zhao, Kaifeng and Zhang, Siwei and Lyu, Xiaozhong and Dusmanu, Mihai and Zhang, Yan and Pollefeys, Marc and Tang, Siyu},
  booktitle={Proceedings of the IEEE/CVF Conference on Computer Vision and Pattern Recognition},
  pages={14497--14509},
  year={2024}
}

@inproceedings{huang2025subjectdrive,
  title={Subjectdrive: Scaling generative data in autonomous driving via subject control},
  author={Huang, Binyuan and Wen, Yuqing and Zhao, Yucheng and Hu, Yaosi and Liu, Yingfei and Jia, Fan and Mao, Weixin and Wang, Tiancai and Zhang, Chi and Chen, Chang Wen and others},
  booktitle={Proceedings of the AAAI Conference on Artificial Intelligence},
  volume={39},
  number={4},
  pages={3617--3625},
  year={2025}
}

@inproceedings{yang2025drivearena,
  title={Drivearena: A closed-loop generative simulation platform for autonomous driving},
  author={Yang, Xuemeng and Wen, Licheng and Wei, Tiantian and Ma, Yukai and Mei, Jianbiao and Li, Xin and Lei, Wenjie and Fu, Daocheng and Cai, Pinlong and Dou, Min and others},
  booktitle={Proceedings of the IEEE/CVF International Conference on Computer Vision},
  pages={26933--26943},
  year={2025}
}

@inproceedings{wang2024robogen,
author = {Wang, Yufei and Xian, Zhou and Chen, Feng and Wang, Tsun-Hsuan and Wang, Yian and Fragkiadaki, Katerina and Erickson, Zackory and Held, David and Gan, Chuang},
title = {RoboGen: towards unleashing infinite data for automated robot learning via generative simulation},
year = {2024},
publisher = {JMLR.org},
booktitle = {Proceedings of the 41st International Conference on Machine Learning},
articleno = {2127},
numpages = {48},
location = {Vienna, Austria},
series = {ICML'24}
}

@misc{chen2025robotwin20scalabledata,
      title={RoboTwin 2.0: A Scalable Data Generator and Benchmark with Strong Domain Randomization for Robust Bimanual Robotic Manipulation}, 
      author={Tianxing Chen and Zanxin Chen and Baijun Chen and Zijian Cai and Yibin Liu and Zixuan Li and Qiwei Liang and Xianliang Lin and Yiheng Ge and Zhenyu Gu and Weiliang Deng and Yubin Guo and Tian Nian and Xuanbing Xie and Qiangyu Chen and Kailun Su and Tianling Xu and Guodong Liu and Mengkang Hu and Huan-ang Gao and Kaixuan Wang and Zhixuan Liang and Yusen Qin and Xiaokang Yang and Ping Luo and Yao Mu},
      year={2025},
      eprint={2506.18088},
      archivePrefix={arXiv},
      primaryClass={cs.RO},
      url={https://arxiv.org/abs/2506.18088}, 
}

@article{singh2024synthetica,
  title={Synthetica: Large Scale Synthetic Data for Robot Perception},
  author={Singh, Ritvik and Liu, Jingzhou and Van Wyk, Karl and Chao, Yu-Wei and Lafleche, Jean-Francois and Shkurti, Florian and Ratliff, Nathan and Handa, Ankur},
  journal={arXiv preprint arXiv:2410.21153},
  year={2024}
}

@ARTICLE{chen2025,
  author={Chen, Yurui and Zhang, Junge and Xie, Ziyang and Li, Wenye and Zhang, Feihu and Lu, Jiachen and Zhang, Li},
  journal={IEEE Transactions on Pattern Analysis and Machine Intelligence}, 
  title={S-NeRF++: Autonomous Driving Simulation via Neural Reconstruction and Generation}, 
  year={2025},
  volume={47},
  number={6},
  pages={4358-4376},
  doi={10.1109/TPAMI.2025.3543072}}

@INPROCEEDINGS{Tang2025,
  author={Tang, Grace and Rajkumar, Swetha and Zhou, Yifei and Walke, Homer Rich and Levine, Sergey and Fang, Kuan},
  booktitle={2025 IEEE International Conference on Robotics and Automation (ICRA)}, 
  title={KALIE: Fine-Tuning Vision-Language Models for Open-World Manipulation Without Robot Data}, 
  year={2025},
  volume={},
  number={},
  pages={9507-9515},
  doi={10.1109/ICRA55743.2025.11128156}}

@ARTICLE{ding2023,
  author={Ding, Wenhao and Xu, Chejian and Arief, Mansur and Lin, Haohong and Li, Bo and Zhao, Ding},
  journal={IEEE Transactions on Intelligent Transportation Systems}, 
  title={A Survey on Safety-Critical Driving Scenario Generation—A Methodological Perspective}, 
  year={2023},
  volume={24},
  number={7},
  pages={6971-6988},
  doi={10.1109/TITS.2023.3259322}}

@article{mandlekar2023mimicgen,
  title={Mimicgen: A data generation system for scalable robot learning using human demonstrations},
  author={Mandlekar, Ajay and Nasiriany, Soroush and Wen, Bowen and Akinola, Iretiayo and Narang, Yashraj and Fan, Linxi and Zhu, Yuke and Fox, Dieter},
  journal={arXiv preprint arXiv:2310.17596},
  year={2023}
}

@article{DEMELO2022174,
title = {Next-generation deep learning based on simulators and synthetic data},
journal = {Trends in Cognitive Sciences},
volume = {26},
number = {2},
pages = {174-187},
year = {2022},
issn = {1364-6613},
doi = {https://doi.org/10.1016/j.tics.2021.11.008},
url = {https://www.sciencedirect.com/science/article/pii/S136466132100293X},
author = {Celso M. {de Melo} and Antonio Torralba and Leonidas Guibas and James DiCarlo and Rama Chellappa and Jessica Hodgins}
}

@article{ren2025cosmos,
  title={Cosmos-Drive-Dreams: Scalable Synthetic Driving Data Generation with World Foundation Models},
  author={Ren, Xuanchi and Lu, Yifan and Cao, Tianshi and Gao, Ruiyuan and Huang, Shengyu and Sabour, Amirmojtaba and Shen, Tianchang and Pfaff, Tobias and Wu, Jay Zhangjie and Chen, Runjian and others},
  journal={arXiv preprint arXiv:2506.09042},
  year={2025}
}

@article{gao2024survey,
  title={A survey for foundation models in autonomous driving},
  author={Gao, Haoxiang and Wang, Zhongruo and Li, Yaqian and Long, Kaiwen and Yang, Ming and Shen, Yiqing},
  journal={arXiv preprint arXiv:2402.01105},
  year={2024}
}

@ARTICLE{11098567,
  author={Liu, Yang and Chen, Weixing and Bai, Yongjie and Liang, Xiaodan and Li, Guanbin and Gao, Wen and Lin, Liang},
  journal={IEEE/ASME Transactions on Mechatronics}, 
  title={Aligning Cyber Space With Physical World: A Comprehensive Survey on Embodied {AI}}, 
  year={2025},
  volume={30},
  number={6},
  pages={7253-7274},
  doi={10.1109/TMECH.2025.3574943}}

@article{sui2025grounding,
  title={From Grounding to Manipulation: Case Studies of Foundation Model Integration in Embodied Robotic Systems},
  author={Sui, Xiuchao and Tian, Daiying and Sun, Qi and Chen, Ruirui and Choi, Dongkyu and Kwok, Kenneth and Poria, Soujanya},
  journal={arXiv preprint arXiv:2505.15685},
  year={2025}
}

@inproceedings{zheng2025universal,
  title={Universal actions for enhanced embodied foundation models},
  author={Zheng, Jinliang and Li, Jianxiong and Liu, Dongxiu and Zheng, Yinan and Wang, Zhihao and Ou, Zhonghong and Liu, Yu and Liu, Jingjing and Zhang, Ya-Qin and Zhan, Xianyuan},
  booktitle={Proceedings of the Computer Vision and Pattern Recognition Conference},
  pages={22508--22519},
  year={2025}
}

@article{deng2025graspvla,
  title={Graspvla: a grasping foundation model pre-trained on billion-scale synthetic action data},
  author={Deng, Shengliang and Yan, Mi and Wei, Songlin and Ma, Haixin and Yang, Yuxin and Chen, Jiayi and Zhang, Zhiqi and Yang, Taoyu and Zhang, Xuheng and Zhang, Wenhao and others},
  journal={arXiv preprint arXiv:2505.03233},
  year={2025}
}

@article{ahn2024autort,
  title={{AutoRT}: Embodied foundation models for large scale orchestration of robotic agents},
  author={Ahn, Michael and Dwibedi, Debidatta and Finn, Chelsea and Arenas, Montse Gonzalez and Gopalakrishnan, Keerthana and Hausman, Karol and Ichter, Brian and Irpan, Alex and Joshi, Nikhil and Julian, Ryan and others},
  journal={arXiv preprint arXiv:2401.12963},
  year={2024}
}

@article{hu2022lora,
  title={{LoRA}: Low-rank adaptation of large language models.},
  author={Hu, Edward J and Shen, Yelong and Wallis, Phillip and Allen-Zhu, Zeyuan and Li, Yuanzhi and Wang, Shean and Wang, Lu and Chen, Weizhu and others},
  journal={ICLR},
  volume={1},
  number={2},
  pages={3},
  year={2022}
}

@article{cross2024hypothetical,
  title={Hypothetical minds: Scaffolding theory of mind for multi-agent tasks with large language models},
  author={Cross, Logan and Xiang, Violet and Bhatia, Agam and Yamins, Daniel LK and Haber, Nick},
  journal={arXiv preprint arXiv:2407.07086},
  year={2024}
}

@inproceedings{devlin2019bert,
  title={Bert: Pre-training of deep bidirectional transformers for language understanding},
  author={Devlin, Jacob and Chang, Ming-Wei and Lee, Kenton and Toutanova, Kristina},
  booktitle={Proceedings of the 2019 conference of the North American chapter of the association for computational linguistics: human language technologies, volume 1 (long and short papers)},
  pages={4171--4186},
  year={2019}
}

@inproceedings{liu2024heirarchical,
author = {Liu, Jijia and Yu, Chao and Gao, Jiaxuan and Xie, Yuqing and Liao, Qingmin and Wu, Yi and Wang, Yu},
title = {LLM-Powered Hierarchical Language Agent for Real-time Human-AI Coordination},
year = {2024},
isbn = {9798400704864},
publisher = {International Foundation for Autonomous Agents and Multiagent Systems},
address = {Richland, SC},
booktitle = {Proceedings of the 23rd International Conference on Autonomous Agents and Multiagent Systems},
pages = {1219–1228},
numpages = {10},
location = {Auckland, New Zealand},
series = {AAMAS '24}
}

@book {dhakal2020cognitive,
	Title = {Cognitive Deficits},
	Author = {Dhakal, Aayush and Bobrin, Bradford D.},
	Publisher = {StatPearls Publishing, Treasure Island (FL)},
	Year = {2025},
	URL = {http://europepmc.org/books/NBK559052},
}

@article{liu2023visual,
  title={Visual instruction tuning},
  author={Liu, Haotian and Li, Chunyuan and Wu, Qingyang and Lee, Yong Jae},
  journal={Advances in neural information processing systems},
  volume={36},
  pages={34892--34916},
  year={2023}
}

@article{yun2025sleepless,
  title={Sleepless nights, sugary days: Creating synthetic users with health conditions for realistic coaching agent interactions},
  author={Yun, Taedong and Yang, Eric and Safdari, Mustafa and Lee, Jong Ha and Kumar, Vaishnavi Vinod and Mahdavi, S Sara and Amar, Jonathan and Peyton, Derek and Aharony, Reut and Michaelides, Andreas and others},
  journal={arXiv preprint arXiv:2502.13135},
  year={2025}
}

@article{decastro2024dreaming,
  title={Dreaming to Assist: Learning to Align with Human Objectives for Shared Control in High-Speed Racing},
  author={DeCastro, Jonathan and Silva, Andrew and Gopinath, Deepak and Sumner, Emily and Balch, Thomas M and Dees, Laporsha and Rosman, Guy},
  journal={arXiv preprint arXiv:2410.10062},
  year={2024}
}

@article{yang2024trajectory,
  title={Trajectory improvement and reward learning from comparative language feedback},
  author={Yang, Zhaojing and Jun, Miru and Tien, Jeremy and Russell, Stuart J and Dragan, Anca and B{\i}y{\i}k, Erdem},
  journal={arXiv preprint arXiv:2410.06401},
  year={2024}
}

@inproceedings{verghese2025user,
  title={User-in-the-loop Evaluation of Multimodal LLMs for Activity Assistance},
  author={Verghese, Mrinal and Chen, Brian and Eghbalzadeh, Hamid and Nagarajan, Tushar and Desai, Ruta},
  booktitle={2025 IEEE/CVF Winter Conference on Applications of Computer Vision (WACV)},
  pages={1144--1154},
  year={2025},
  organization={IEEE}
}

@article{gopinath2024computational,
  title={Computational Teaching for Driving via Multi-Task Imitation Learning},
  author={Gopinath, Deepak and Cui, Xiongyi and DeCastro, Jonathan and Sumner, Emily and Costa, Jean and Yasuda, Hiroshi and Morgan, Allison and Dees, Laporsha and Chau, Sheryl and Leonard, John and others},
  journal={arXiv preprint arXiv:2410.01608},
  year={2024}
}

@article{liu2023reflect,
  title={Reflect: Summarizing robot experiences for failure explanation and correction},
  author={Liu, Zeyi and Bahety, Arpit and Song, Shuran},
  journal={arXiv preprint arXiv:2306.15724},
  year={2023}
}

@article{guan2024task,
  title={Task success is not enough: Investigating the use of video-language models as behavior critics for catching undesirable agent behaviors},
  author={Guan, Lin and Zhou, Yifan and Liu, Denis and Zha, Yantian and Amor, Heni Ben and Kambhampati, Subbarao},
  journal={arXiv preprint arXiv:2402.04210},
  year={2024}
}

@article{strouse2021collaborating,
  title={Collaborating with humans without human data},
  author={Strouse, DJ and McKee, Kevin and Botvinick, Matt and Hughes, Edward and Everett, Richard},
  journal={Advances in Neural Information Processing Systems},
  volume={34},
  pages={14502--14515},
  year={2021}
}

@article{yan2023efficient,
  title={An efficient end-to-end training approach for zero-shot human-AI coordination},
  author={Yan, Xue and Guo, Jiaxian and Lou, Xingzhou and Wang, Jun and Zhang, Haifeng and Du, Yali},
  journal={Advances in Neural Information Processing Systems},
  volume={36},
  pages={2636--2658},
  year={2023}
}

@inproceedings{zhang2024proagent,
  title={Proagent: building proactive cooperative agents with large language models},
  author={Zhang, Ceyao and Yang, Kaijie and Hu, Siyi and Wang, Zihao and Li, Guanghe and Sun, Yihang and Zhang, Cheng and Zhang, Zhaowei and Liu, Anji and Zhu, Song-Chun and others},
  booktitle={Proceedings of the AAAI Conference on Artificial Intelligence},
  volume={38},
  number={16},
  pages={17591--17599},
  year={2024}
}

@article{pearce2023imitating,
  title={Imitating human behaviour with diffusion models},
  author={Pearce, Tim and Rashid, Tabish and Kanervisto, Anssi and Bignell, Dave and Sun, Mingfei and Georgescu, Raluca and Macua, Sergio Valcarcel and Tan, Shan Zheng and Momennejad, Ida and Hofmann, Katja and others},
  journal={arXiv preprint arXiv:2301.10677},
  year={2023}
}

@article{laidlaw2022boltzmann,
  title={The boltzmann policy distribution: Accounting for systematic suboptimality in human models},
  author={Laidlaw, Cassidy and Dragan, Anca},
  journal={arXiv preprint arXiv:2204.10759},
  year={2022}
}

@INPROCEEDINGS{zhang2023large,

  author={Zhang, Bowen and Soh, Harold},

  booktitle={2023 IEEE/RSJ International Conference on Intelligent Robots and Systems (IROS)}, 

  title={Large Language Models as Zero-Shot Human Models for Human-Robot Interaction}, 

  year={2023},

  volume={},

  number={},

  pages={7961-7968},

  doi={10.1109/IROS55552.2023.10341488}}

@inproceedings{wang2023utility,
author = {Wang, Chenxu and Chen, Zilong and Liu, Huaping},
title = {On the Utility of External Agent Intention Predictor for Human-AI Coordination},
year = {2024},
isbn = {9798400704864},
publisher = {International Foundation for Autonomous Agents and Multiagent Systems},
address = {Richland, SC},
booktitle = {Proceedings of the 23rd International Conference on Autonomous Agents and Multiagent Systems},
pages = {2546–2548},
numpages = {3},
location = {Auckland, New Zealand},
series = {AAMAS '24}
}

@article{anthis2025llm,
  title={{LLM} social simulations are a promising research method},
  author={Anthis, Jacy Reese and Liu, Ryan and Richardson, Sean M and Kozlowski, Austin C and Koch, Bernard and Evans, James and Brynjolfsson, Erik and Bernstein, Michael},
  journal={arXiv preprint arXiv:2504.02234},
  year={2025}
}

@article{dosovitskiy2020image,
  title={An image is worth 16x16 words: Transformers for image recognition at scale},
  author={Dosovitskiy, Alexey and Beyer, Lucas and Kolesnikov, Alexander and Weissenborn, Dirk and Zhai, Xiaohua and Unterthiner, Thomas and Dehghani, Mostafa and Minderer, Matthias and Heigold, Georg and Gelly, Sylvain and others},
  journal={arXiv preprint arXiv:2010.11929},
  year={2020}
}

@article{grattafiori2024llama,
  title={The {Llama} 3 herd of models},
  author={Grattafiori, Aaron and Dubey, Abhimanyu and Jauhri, Abhinav and Pandey, Abhinav and Kadian, Abhishek and Al-Dahle, Ahmad and Letman, Aiesha and Mathur, Akhil and Schelten, Alan and Vaughan, Alex and others},
  journal={arXiv preprint arXiv:2407.21783},
  year={2024}
}

@article{ankile2023discovering,
  title={Discovering user types: Mapping user traits by task-specific behaviors in reinforcement learning},
  author={Ankile, Lars L and Ham, Brian S and Mao, Kevin and Shin, Eura and Swaroop, Siddharth and Doshi-Velez, Finale and Pan, Weiwei},
  journal={arXiv preprint arXiv:2307.08169},
  year={2023}
}

@inproceedings{silvastability,
  title={Stability of Preference Alignment for Multi-Turn Control with {LLM} Policies},
  author={Silva, Andrew and Tambwekar, Pradyumna and Gopinath, Deepak Edakkattil and DeCastro, Jonathan and Rosman, Guy and Balachandran, Avinash},
  booktitle={First Workshop on Multi-Turn Interactions in Large Language Models},
  year=2025,
}

@article{rajabi2024gsr,
  title={{GSR-BENCH}: A benchmark for grounded spatial reasoning evaluation via multimodal llms},
  author={Rajabi, Navid and Kosecka, Jana},
  journal={arXiv preprint arXiv:2406.13246},
  year={2024}
}

@article{zhang2025mllms,
  title={Why do {MLLMs} struggle with spatial understanding? a systematic analysis from data to architecture},
  author={Zhang, Wanyue and Huang, Yibin and Xu, Yangbin and Huang, JingJing and Zhi, Helu and Ren, Shuo and Xu, Wang and Zhang, Jiajun},
  journal={arXiv preprint arXiv:2509.02359},
  year={2025}
}

@article{ouyang2022training,
  title={Training language models to follow instructions with human feedback},
  author={Ouyang, Long and Wu, Jeffrey and Jiang, Xu and Almeida, Diogo and Wainwright, Carroll and Mishkin, Pamela and Zhang, Chong and Agarwal, Sandhini and Slama, Katarina and Ray, Alex and others},
  journal={Advances in neural information processing systems},
  volume={35},
  pages={27730--27744},
  year={2022}
}

@article{boyle2025robotxr1,
  title={RobotxR1: Enabling Embodied Robotic Intelligence on Large Language Models through Closed-Loop Reinforcement Learning},
  author={Boyle, Liam and Baumann, Nicolas and Sivasothilingam, Paviththiren and Magno, Michele and Benini, Luca},
  journal={arXiv preprint arXiv:2505.03238},
  year={2025}
}

@article{ding2025empowering,
  title={Empowering Multi-Turn Tool-Integrated Reasoning with Group Turn Policy Optimization},
  author={Ding, Yifeng and Le, Hung and Han, Songyang and Ruan, Kangrui and Jin, Zhenghui and Kumar, Varun and Wang, Zijian and Deoras, Anoop},
  journal={arXiv preprint arXiv:2511.14846},
  year={2025}
}

@inproceedings{zhao2022coordination,
  title={Coordination with humans via strategy matching},
  author={Zhao, Michelle and Simmons, Reid and Admoni, Henny},
  booktitle={International Conference on Intelligent Robots and Systems (IROS)},
  pages={9116--9123},
  year={2022},
  organization={IEEE}
}

@article{carroll2019utility,
  title={On the utility of learning about humans for human-ai coordination},
  author={Carroll, Micah and Shah, Rohin and Ho, Mark K and Griffiths, Tom and Seshia, Sanjit and Abbeel, Pieter and Dragan, Anca},
  journal={Advances in neural information processing systems},
  volume={32},
  year={2019}
}

@article{zapata2025advancements,
  title={Advancements in Assistive Robotics: A Systematic Review of Inclusive Technologies for People With Disabilities},
  author={Zapata, Mireya and Guevara, Diego and Obreg{\'o}n, Jennifer and Arias-Flores, Hugo and Ramos-Galarza, Carlos},
  journal={IEEE Access},
  volume={13},
  pages={171871--171888},
  year={2025}
}

@article{hoffman2024inferring,
  title={Inferring human intent and predicting human action in human--robot collaboration},
  author={Hoffman, Guy and Bhattacharjee, Tapomayukh and Nikolaidis, Stefanos},
  journal={Annual Review of Control, Robotics, and Autonomous Systems},
  volume={7},
  year={2024},
  publisher={Annual Reviews}
}

\appendix
\section{Experimental Setup}
\label{sec:A-exp-setup}

The models which used the 1 billion parameter Llama3 models were trained on either a g6e.12xlarge instance on AWS or a local cluster with four Nvidia RTX 6000 (49 GB memory). The models which used the 8 billion parameter Llama3 models were trained on the p5.48xlarge AWS instance, as these models utilized over 80 GB of GPU-memory.  
Both the visual encoder, and language model are trained using Low-Rank Adaptation (LoRA)~\citep{hu2022lora}. 
The multimodal projection layer was trained from scratch.
The specific values for our experimental parameters are provide in Table~\ref{tab:exp-params}.

\begin{table}[ht]
\centering
\caption{Experimental parameters per baseline. All used image + llm LoRA with r, alpha = 16. Image -> LLM projection layer trained from scratch. 500 warmup steps. }
\label{tab:exp-params}
\small
\begin{tabular}{cccccccc}
\toprule
Model-name &
\shortstack{Training\\Datasets}&
Data Aug &
Learning rate &
\shortstack{Training\\Steps} &
\shortstack{Per-device\\ batch size} &
\shortstack{Weight\\Decay} \\
\midrule
Ours (1B) & $D_{train}$  & No  & $5e-5$  & 15000 & 1 & 0.01 \\
\shortstack{1B + grounding datasets}  & $\mathcal{D}_{train} + \mathcal{D}_{ground}$  & Yes & $5e-5$ & 16500 & 1 & 0.01 \\
1B Coaching Only         & $\mathcal{D}^{coach}$ & Yes & $5e-5$  & 5000 & 1 & 0.01 \\
1B Correction Only         & $\mathcal{D}^{correct}$ & No  & $5e-5$  & 5000 & 1 & 0.01 \\
Ours (1B) + I-QA        & $\mathcal{D}_{train} + \mathcal{D}^I$ & Yes & $5e-5$  & 16500 & 1 & 0.01 \\
Ours (1B) + T-QA        & $\mathcal{D}_{train} + \mathcal{D}^T$ & Yes & $5e-5$ & 16500 & 1 & 0.01 \\
Ours (1B) + V-QA        & $\mathcal{D}_{train} + \mathcal{D}^V$ & Yes & $5e-5$ & 16500 & 1 & 0.01 \\
\bottomrule
\end{tabular}
\end{table}

\begin{figure}[htbp]
    \centering
    \includegraphics[width=\textwidth]{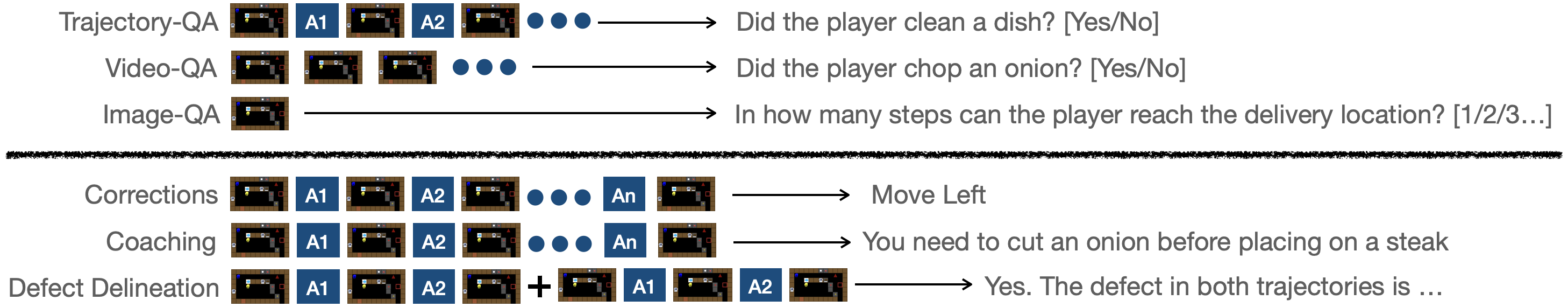}
    \caption{This figure provides an abstract depiction of each task synthesized in this paper to train our assistive model. Top to Bottom - Trajectory-QA, Video-QA, Image-QA, Corrections, Coaching, Defect Delineation}
    \label{fig:dataset_diagram}
\end{figure}

\begin{figure}[htbp]
    \centering
    \includegraphics[width=\textwidth]{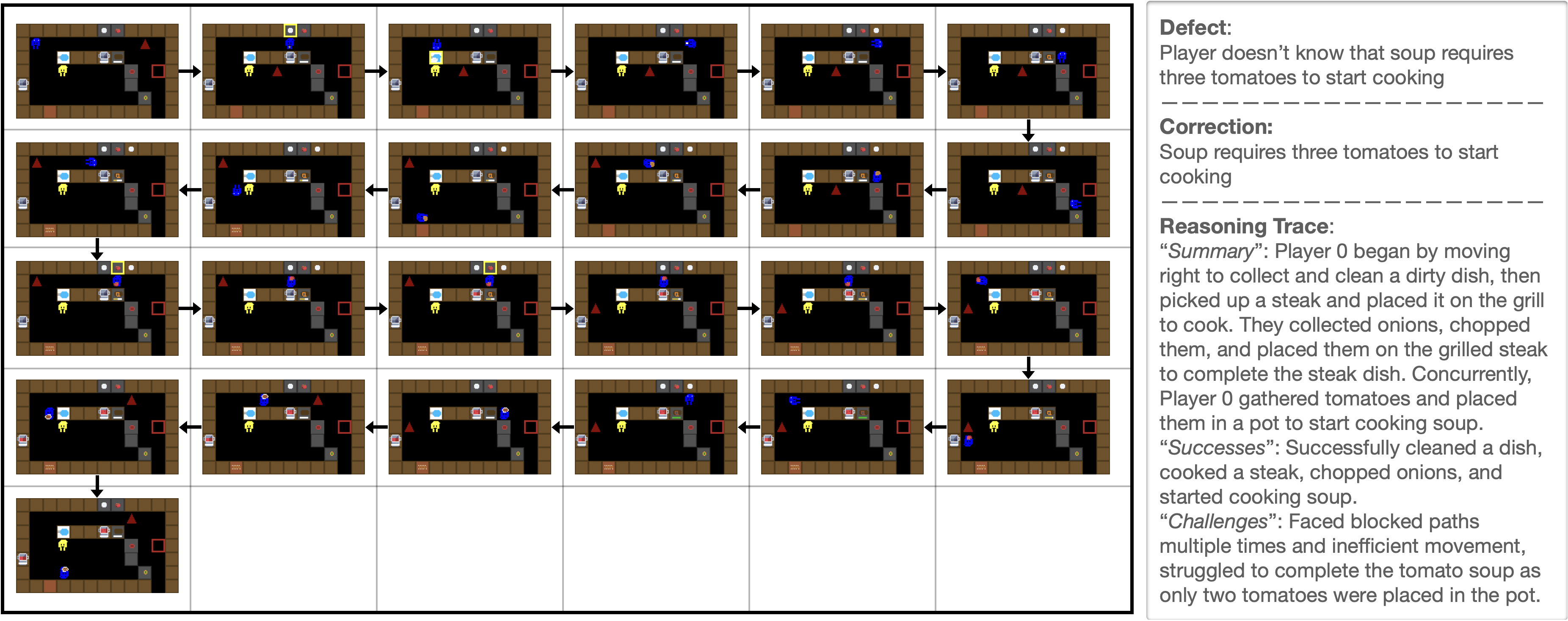}
    \caption{Dataset Example \color{black}}
    \label{fig:coaching_example}
\end{figure}

\begin{figure}[htbp]
    \centering
    \includegraphics[width=\textwidth]{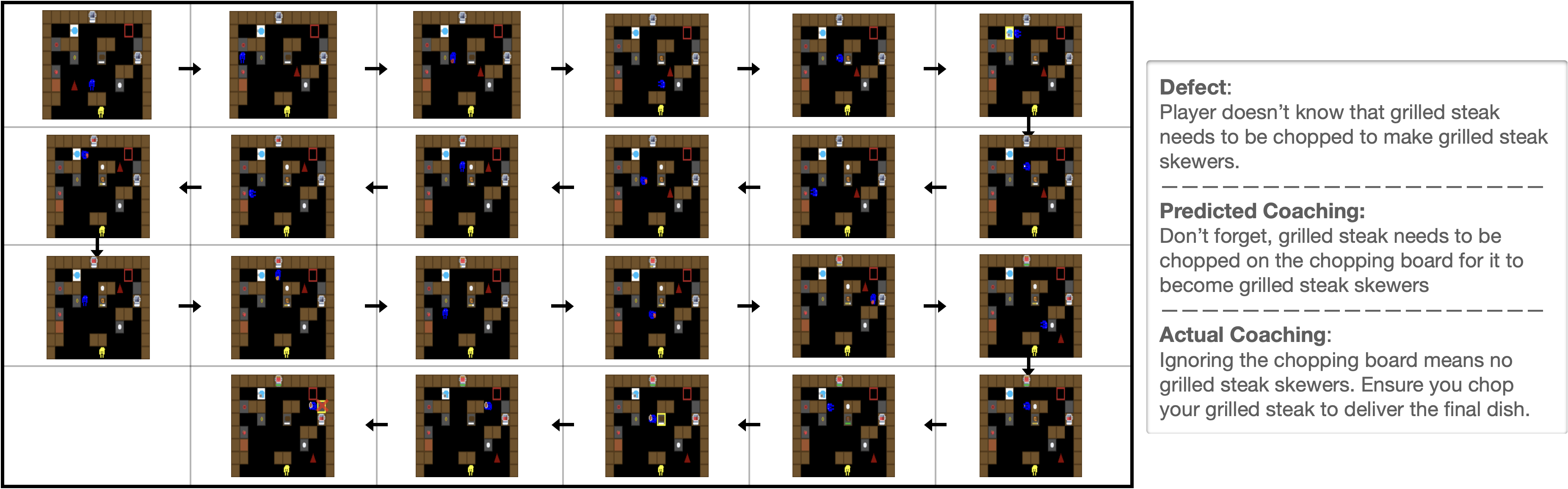}
    \caption{Model Output \color{black}}
    \label{fig:overview}
\end{figure}

\section{Prompts}
\subsection{Map Generation}
\label{sec:map-prompt}
We synthetically generated maps to rollout our defective heuristics and obtain diverse behavior trajectories (see Figure~\ref{fig:map-gen-prompt}). 
Each map was checked to ensure satsifaction with common-sense and validity checks to ensure that they yield reasonable trajectories. 

\begin{center}

  % --- SYSTEM (breakable) ---
  \begin{tcolorbox}[promptbox, breakable,
    title={System: Map Generation},
    colbacktitle=systemc!90!black]
\begin{Verbatim}[fontsize=\footnotesize]
I need you to create new overcooked maps. I will provide you with a sample map. You need to generate new maps, with the same number of each object. 
    
Here is a dictionary which identifies each symbol in the map: \n {map_key}

Generate a map while keeping the following things in mind: 
(1) Make sure there is at least one of each of the following symbols, ["O", "D", "T", "C", "N", "S", "H", "G", "I"] 
(2) Make sure that Player 0 can access all target locations, ["O", "D", "T", "C", "N", "S", "H", "G", "I"]. 
(3) You are allowed to change the size of the outer grid, as long as the number of the "special" locations remain the same. 
(4) Limit the map size to a 12 x 12 grid.  
(5) Make sure you have at least one "X" and one "Y" in the map.
(6) Including spaces, each row should have the same number of characters.
(7) Generate only one map.

Your response should just include the map. 
\end{Verbatim}
  \end{tcolorbox}

  % --- USER (unbreakable) ---
  \begin{tcolorbox}[promptbox, unbreakable,
    title={User: Map Generation},
    colbacktitle=userc!90!black]
\begin{Verbatim}[fontsize=\footnotesize]
Here are a few examples of a few possible overcooked map-layouts:

Map 6: 
\end{Verbatim}
  \end{tcolorbox}

  \captionof{figure}{Prompt utilized to enable GPT-4o to generate overcooked Map configurations.}
  \label{fig:map-gen-prompt}
\end{center}

\subsection{Coaching}
\label{sec:coach-gen}
We leveraged GPT-4o to synthetically generate language-based corrective feedback for each trajectory, to teach our model to coach a synthetic user with defective behavior. The process of generating coaching snippets was two-fold: (1) We first generated a set of initial coaching snippets given a description of the defect (see Figure~\ref{fig:coaching-gen-prompt}), (2) Next, we augmented these coaching snippets using a set of directives such as ``Make the advice more concise'' or ``Make the advice more humourous.'' (see Figure~\ref{fig:coaching-gen-prompt-2})

\begin{center}

  % --- SYSTEM (breakable) ---
  \begin{tcolorbox}[promptbox, breakable,
    title={System Prompt},
    colbacktitle=systemc!90!black]
\begin{Verbatim}[fontsize=\footnotesize]
I need to generate corrective feedback to enable a user to fix a specific defect in their strategy for the game overcooked. I will provide you a description of the defect and you need to provide some options for how you would assist a user in correcting this defect . Please craft the exact statement you would say to a user.

Each correction should be brief (maximum of two sentences). You should address only the specific defect mentioned, avoiding general advice or including irrelevant information. For example, discuss grills only if the defect pertains to grills, if you are asked to correct a defect pertaining to general interactions with dispensers do not specifically mention any individual dispenser (onion, dish, tomato, steak, etc.).
    
Note: Make sure to keep your feedback grounded in overcooked rather than an actual kitchen. Make sure your corrections are consistent with the rules of overcooked: {rules}
\end{Verbatim}
  \end{tcolorbox}

  % --- USER (unbreakable; will move to next page if needed) ---
  \begin{tcolorbox}[promptbox, unbreakable,
    title={User Prompt},
    colbacktitle=userc!90!black]
\begin{Verbatim}[fontsize=\footnotesize]
Please give me five different potential corrections for the following defect. Your corrections should be in the following format: 
     
    Correction1: ....
    Correction2: ....
    Correction3: ....
    Correction4: ....
    Correction5: ....

    Defect: {defect}
\end{Verbatim}
  \end{tcolorbox}

  \captionof{figure}{Prompts used for generating initial coaching snippet from the defect}
  \label{fig:coaching-gen-prompt}
\end{center}

\begin{center}

  % --- SYSTEM (breakable) ---
  \begin{tcolorbox}[promptbox, breakable,
    title={System Prompt},
    colbacktitle=systemc!90!black]
\begin{Verbatim}[fontsize=\footnotesize]
I have a set of corrections to help a user fix a defect in their gameplay strategy.Your job is to rephrase this set of corrections according to a directive I will provide to you. The instructions will be things like "make the correction more polite," "make the correction more direct," "make the correction more urgent," etc. However, you must ensure that the meaning of the correction should still remain the same, i.e., the correction should still be helpful in correcting the given defect in the user's strategy.
Make sure that your augmented corrections are still grounded in overcooked and not in an actual kitchen setting. Try not to add unnecessary filler words if you can avoid it. The input will be provided to you in the following format:

    ## Player Defect
    Defect: <description of the defect>
    
    ## Initial Corrections
    Correction1: <correction1>
    Correction2: <correction2>
    Correction3: <correction3>
    Correction4: <correction4>
    Correction5: <correction5>

    ## Directive - <directive>
     
    ## Augmented Corrections (TODO)
    Augmented Correction1: <updated correction1>
    Augmented Correction2: <updated correction2>
    Augmented Correction3: <updated correction3>
    Augmented Correction4: <updated correction4>
    Augmented Correction5: <updated correction5> 

Make sure your corrections are consistent with the rules of overcooked: {rules}
\end{Verbatim}
  \end{tcolorbox}

  % --- USER (unbreakable) ---
  \begin{tcolorbox}[promptbox, unbreakable,
    title={User Prompt},
    colbacktitle=userc!90!black]
\begin{Verbatim}[fontsize=\small]
Please update the given set of corrections for the following defect using the directive provided. Note that you need to augment each correction individually according to the directive, i.e., you are generating one augmented correction for every correction you have been provided. --  

    ## Player Defect
    Defect: {defect}

    ## Initial Corrections
    Correction1: {corrections[0]}
    Correction2: {corrections[1]}
    Correction3: {corrections[2]}
    Correction4: {corrections[3]}
    Correction5: {corrections[4]}

    ## Directive - {directive}
\end{Verbatim}
  \end{tcolorbox}

  \captionof{figure}{Prompts used for augmenting initial coaching snippets based on a set of directives.}
  \label{fig:coaching-gen-prompt-2}
\end{center}

\subsection{Reasoning Generation}
\label{sec:reasoning-gen-prompt}
We prompt GPT-4o to generate reasoning traces by using the list of events which occurred in a trajectory (see Figure~\ref{fig:reasoning-gen-prompt}). 
Each reasoning trace is comprised of three components; (1) Summary, (2) Successes, (3) Challenges. 
This information is included in the ``format instructions'' portion ofthe prompt. 
The model is also provided with the list of rules in overcooked before reasoning over the trajectory. 
The specific set of rules included in the prompt vary based on the recipe the user is attempting to make (see Figures~\ref{fig:rules},~\ref{fig:rules-1},~\ref{fig:rules-2},~\ref{fig:rules-3}).

\begin{center}

  % --- SYSTEM (breakable) ---
  \begin{tcolorbox}[promptbox, breakable,
    title={Reasoning Trace Generation Prompt},
    colbacktitle=systemc!90!black]
\begin{Verbatim}[fontsize=\footnotesize]
You are an expert game analyst. Ground your analysis in the following rules of Overcooked:{overcooked_rules}"

Generate a structured summary of an Overcooked gameplay trajectory with the following format requirements: {format_instructions}
After summarizing, for each field perform self-verification: indicate if it's supported by the events and rules ('valid') and a confidence score between 0-1, under 'verification'

Respond strictly with the JSON object, no extra text.
Events (chronological list): {events}
\end{Verbatim}
  \end{tcolorbox}

  \captionof{figure}{Reasoning Generation Prompt}
  \label{fig:reasoning-gen-prompt}
\end{center}

\begin{center}

  % --- SYSTEM (breakable) ---
  \begin{tcolorbox}[promptbox, breakable,
    title={Overcooked Rules (original recipe)},
    colbacktitle=systemc!90!black]
\begin{Verbatim}[fontsize=\footnotesize]
Overcooked Rules: 
1. Players must pick up ingredients from dispensers and place them in the cooking ranges to cook them.
2. Tomatoes are placed in the cooking pots to make soup.
3. Players must interact with dispensers to pick up ingredients.
4. Onions need to be cut at the cutting board.
5. Steak needs to be cooked on the grill.
6. There is one visible trip hazard and one invisible trip hazard on the map.
7. If the player walks over a hazard while holding an object, they may drop the object.
8. Players must deliver cooked food to the delivery location.
9. A player can only hold one object at a time. 
10. Dishes are cleaned at the sink.
11. A player can perform an action at a location by interacting with the location, i.e. cleaning dishes at the sink, or cutting onions at the cutting board.
12. Soup starts cooking when there are three tomatoes in the pot. 
13. Steak starts cooking when there is one steak on the grill.
14. Once food is cooking, you can no longer place ingredients on it until the food is cooked and picked up.  
15. If you want to drop an ingredient, you have to drop it on a counter. 
16. You are only controlling one player in overcooked.
17. Interacting with cooking ranges before the food is cooked will not do anything.
18. Cut onions can only be picked up once you are holding grilled steak.
19. The only ingredients in the game are steak, tomatoes, and onions. 
20. The game has no audio, i.e. the player only receives visual ques while playing the game. 
\end{Verbatim}
  \end{tcolorbox}
  
  \captionof{figure}{Overcooked rules}
  \label{fig:rules}
\end{center}

\begin{center}

  % --- SYSTEM (breakable) ---
  \begin{tcolorbox}[promptbox, breakable,
    title={Overcooked Rules (alternate1 recipe)},
    colbacktitle=systemc!90!black]
\begin{Verbatim}[fontsize=\footnotesize]
Overcooked Rules: 
1. Players must pick up ingredients from dispensers and place them in the cooking ranges to cook them.
2. Tomatoes are placed in the cooking pots to make soup.
3. Players must interact with dispensers to pick up ingredients.
4. Onions need to be cut at the cutting board.
5. Steak needs to be cooked on the grill.
6. There is one visible trip hazard and one invisible trip hazard on the map.
7. If the player walks over a hazard while holding an object, they may drop the object.
8. Players must deliver cooked food to the delivery location.
9. A player can only hold one object at a time. 
10. Dishes are cleaned at the sink.
11. A player can perform an action at a location by interacting with the location, i.e. cleaning dishes at the sink, or cutting onions at the cutting board.
12. Soup starts cooking when there are three tomatoes in the pot. 
13. Steak starts cooking when there is one steak on the grill.
14. The two recipes you need to make are soup with chopped onions and a grilled steak.  
15. If you want to drop an ingredient, you have to drop it on a counter. 
16. You are only controlling one player in overcooked.
17. Interacting with cooking ranges before the food is cooked will not do anything.
18. Chopped onions can only be picked up once you are holding soup.
19. The only ingredients in the game are steak, tomatoes, and onions. 
20. The game has no audio, i.e. the player only receives visual ques while playing the game. 
21. Chopped onions must be picked up from the chopping board while holding soup. 
\end{Verbatim}
  \end{tcolorbox}
  
  \captionof{figure}{Overcooked rules (Alternate 1)}
  \label{fig:rules-1}
\end{center}

\begin{center}

  % --- SYSTEM (breakable) ---
  \begin{tcolorbox}[promptbox, breakable,
    title={Overcooked Rules (alternate2 recipe)},
    colbacktitle=systemc!90!black]
\begin{Verbatim}[fontsize=\footnotesize]
Overcooked Rules: 
1. Players must pick up ingredients from dispensers and place them in the cooking ranges to cook them.
2. Tomatoes are placed in the cooking pots to make soup.
3. Players must interact with dispensers to pick up ingredients.
4. One onion needs to be placed in the cooking pot to make tomato and onion soup.
5. Steak needs to be cooked on the grill.
6. There is one visible trip hazard and one invisible trip hazard on the map.
7. If the player walks over a hazard while holding an object, they may drop the object.
8. Players must deliver cooked food to the delivery location.
9. A player can only hold one object at a time. 
10. Dishes are cleaned at the sink.
11. A player can perform an action at a location by interacting with the location, i.e. cleaning dishes at the sink, or cutting onions at the cutting board.
12. Soup starts cooking when there are two tomatoes and one onion in the cooking pot. 
13. Steak starts cooking when there is one steak on the grill.
14. The only recipe you need to make is tomato, onion and steak stew.  
15. If you want to drop an ingredient, you have to drop it on a counter. 
16. You are only controlling one player in overcooked.
17. Interacting with cooking ranges before the food is cooked will not do anything.
18. You can only place an onion in the cooking pot, once there are two tomatoes in the pot. 
19. The only ingredients in the game are steak, tomatoes, and onions. 
20. The game has no audio, i.e. the player only receives visual ques while playing the game.
21. You need to chop grilled steak on the chopping board. 
22. To complete your stew, you first need to pick up the cooked tomato and onion soup from the pot, and then pick up chopped steak from the chopping board while holding the soup.
\end{Verbatim}
  \end{tcolorbox}
  
  \captionof{figure}{Overcooked rules (Alternate 2)}
  \label{fig:rules-2}
\end{center}

\begin{center}

  % --- SYSTEM (breakable) ---
  \begin{tcolorbox}[promptbox, breakable,
    title={Overcooked Rules (alternate2 recipe)},
    colbacktitle=systemc!90!black]
\begin{Verbatim}[fontsize=\footnotesize]
Overcooked Rules: 
1. Players must pick up ingredients from dispensers and place them in the cooking ranges to cook them.
2. Tomatoes are placed in the cooking pots to make soup.
3. Players must interact with dispensers to pick up ingredients.
4. One onion needs to be placed in the cooking pot to make tomato and onion soup.
5. Steak needs to be cooked on the grill.
6. There is one visible trip hazard and one invisible trip hazard on the map.
7. If the player walks over a hazard while holding an object, they may drop the object.
8. Players must deliver cooked food to the delivery location.
9. A player can only hold one object at a time. 
10. Dishes are cleaned at the sink.
11. A player can perform an action at a location by interacting with the location, i.e. cleaning dishes at the sink, or cutting onions at the cutting board.
12. Soup starts cooking when there are two tomatoes and one onion in the cooking pot. 
13. Steak starts cooking when there is one steak on the grill.
14. The two recipes you need to make are tomato and onion soup and grilled steak skewers.  
15. If you want to drop an ingredient, you have to drop it on a counter. 
16. You are only controlling one player in overcooked.
17. Interacting with cooking ranges before the food is cooked will not do anything.
18. You can only place an onion in the cooking pot, once there are two tomatoes in the pot to make tomato and onion soup. 
19. The only ingredients in the game are steak, tomatoes, and onions. 
20. The game has no audio, i.e. the player only receives visual ques while playing the game.
21. You need to chop grilled steak on the chopping board to make grilled steak skewers. 
22. Once you have chopped the grilled steak, you can pick it up and serve the grilled steak skewers.
23. Ingredients cannot be removed once they are placed in the cooking pot or on the grill.
\end{Verbatim}
  \end{tcolorbox}
  
  \captionof{figure}{Overcooked rules (Alternate 3)}
  \label{fig:rules-3}
\end{center}

\subsection{LLM-as-judge}
\label{sec:llm-as-judge}
We prompt GPT-4 to serve as an evaluator to judge the accuracy of the coaching trace generated by our model in comparison to the actual underlying defect in the trajectory (see Figure~\ref{fig:llm-judge}). 

\begin{center}

  % --- SYSTEM (breakable) ---
  \begin{tcolorbox}[promptbox, breakable,
    title={System: LLM-Judge},
    colbacktitle=systemc!90!black]
\begin{Verbatim}[fontsize=\footnotesize]
You are an evaluator reviewing whether coaching advice generated by an AI assistant correctly addresses a specific gameplay defect in Overcooked.
### Task
You are given the following:

- **Generated Advice**: The assistant's response to a gameplay issue.
- **Actual Defect**: The player's observed gameplay problem.
- **Reference Coaching Advice**: A human-written or high-quality model coaching response meant to guide users with this type of defect.

Your job is to decide whether the **Generated Advice** specifically and adequately addresses the **Actual Defect**.

The **Reference Coaching Advice** is provided as helpful context, to better understand the defect being resolved. But the **generated advice** does not need to match it exactly.
---
### Instructions
- Output **Label: [Yes]** if the generated advice *directly and sufficiently addresses* the defect.
- Output **Label: [No]** if it *fails to address*, *only partially addresses*, or is *too vague, off-topic, or incorrect* in relation to the defect.
- Include a short **Think** step to reason through your decision.
---
### Format
### User  
Generated Advice: <generated_answer>  
Actual Defect: <actual_defect>  
Reference Coaching Advice: <reference_advice>  

### Assistant  
Think: <your brief analysis of the advice and defect relationship>  
Label: [Yes/No]  
---
### Examples

#### Example 1
### User  
Generated Advice: Make sure you place the ingredients in the cooking range to start
Actual Defect: The player does not know what to do with the ingredients.  
Reference Coaching Advice: Place ingredients in the pot to cook them. Don't leave them on the counters.  

### Assistant  
Think: The advice, though shorter, tells the player what to do with the ingredients, aligning with the defect.  
Label: [Yes]  
---
#### Example 2

### User  
Generated Advice: Avoid inefficient movement.  
Actual Defect: The player makes unnecessary movements/turns while navigating the map.  
Reference Coaching Advice: Stick to efficient paths and avoid zig-zagging or turning back.  

### Assistant  
Think: The generated advice is high-level but directly relevant to the defect.  
Label: [Yes]  
---
#### Example 3
### User  
Generated Advice: Remember to place three tomatoes in the pot to cook the soup.  
Actual Defect: The player thinks it is best to only serve soup.  
Reference Coaching Advice: It would be helpful to serve both soup and steak, rather than soup alone.  

### Assistant  
Think: The generated advice focuses on preparing soup, which does not address the misconception about ignoring other dishes. The reference advice suggests serving both soup and steak, which is not mentioned in the generated advice. 
Label: [No]  
---
#### Example 4
### User  
Generated Advice: Place three tomatoes in the pot to start the cooking process.  
Actual Defect: The player tries to deliver cooked food by placing it on a counter.  
Reference Coaching Advice: Deliver completed dishes by placing them at the delivery location, not counters.  

### Assistant  
Think: The advice is unrelated to the problem of delivering food to the right location.  
Label: [No]  
---
#### Example 5
### User  
Generated Advice: Remember to slice onions on the board.  
Actual Defect: The player tries to place onions in the sink.  
Reference Coaching Advice: Use the cutting board for prepping onions before cooking. 

### Assistant  
Think: The advice provides the correct instruction that addresses the mistake with the handling of onions.  
Label: [Yes]  
\end{Verbatim}
  \end{tcolorbox}

  % --- USER (unbreakable) ---
  \begin{tcolorbox}[promptbox, unbreakable,
    title={User: LLM-Judge},
    colbacktitle=userc!90!black]
\begin{Verbatim}[fontsize=\footnotesize]
Generated Advice: <|generated_answer|>
Actual Defect: <|answer|>
Reference Coaching Advice: <|reference_answer|>
\end{Verbatim}
  \end{tcolorbox}

  \captionof{figure}{LLM-as-judge to evaluate coaching outputs}
  \label{fig:llm-judge}
\end{center}

\subsection{Behavior-Critic Baseline}
\label{sec:behavior-critic}
Our baseline behavior-critic leverages GPT-4o to perform coaching or correction based on a downsampled video of the defective trajectory. 
The prompts for both tasks can be found in Figure~\ref{fig:gpt-4o-base-prompt}. 
We also enhanced our behavior critic by including the synthetically generated reasoning traces, as per Section~\ref{sec:reasoning-gen-prompt}. The prompts for this enhanced behavior critic can be found in Figure~\ref{fig:GPT-4o prompts with summary trace}.

\begin{center}

  % --- SYSTEM (breakable) ---
  \begin{tcolorbox}[promptbox, breakable,
    title={Behavior Critic (Coaching)},
    colbacktitle=systemc!90!black]
\begin{Verbatim}[fontsize=\footnotesize]
**Task:**
You will be provided a short video clip (sequence of images) of a human playing Overcooked.
The player may or may not have a defect in their gameplay strategy which is hampering their performance. Your job is to analyze this trajectory and provide corrective feedback to enable the user to understand and fix the defect in their overcooked gameplay.

Please share your analysis of what is occuring in the video with regards to the potentially defective gameplay trajectory, prior to providing your feedback. Note that you need to analyze the actions of the blue player.

**Example Responses (you should follow the same JSON output format):**
{examples}

**Note**
The images are key frames extracted from a video and are presented in the same order as the original footage.
Each player will have at most one defect. The potential defects that the player may have will be one of the following categories: 
{defects}
To help you perform the task, here are some important rules to know about playing Overcooked: 
**{overcooked_rules}
**Now, please look at the following sequence of images carefully.**
{extra}
\end{Verbatim}
  \end{tcolorbox}

  % --- USER (unbreakable) ---
  \begin{tcolorbox}[promptbox, unbreakable,
    title={Corrections},
    colbacktitle=userc!90!black]
\begin{Verbatim}[fontsize=\footnotesize]
**Task:**
You are an AI assistant tasked with improving a user's performance in Overcooked. 
To help you perform the task, here are some important rules to know about playing Overcooked: 
**{overcooked_rules}

Given a defective seed trajectory, provide corrective actions, that the user should perform next, that align with the user's strategy. 
Use the accompanying reference trajectory, a non-defective demonstration on a reference map, to infer their strategy.

**Example Responses (you should follow the same JSON output format):**
{examples}

**Note**
You need to predict the immediate next action the user should take to begin to fix the defect in their behavior. You must choose from one of the following 8 actions: [move forward, move backward, move left, move right, turn left, turn right, interact, no action]
**Now, please look at the following sequence of images carefully.**
{extra}
\end{Verbatim}
  \end{tcolorbox}

  \captionof{figure}{Prompts for GPT-4o baselines without reasoning.}
  \label{fig:gpt-4o-base-prompt}
\end{center}

\begin{center}

  % --- SYSTEM (breakable) ---
  \begin{tcolorbox}[promptbox, breakable,
    title={Coaching Prompt},
    colbacktitle=systemc!90!black]
\begin{Verbatim}[fontsize=\footnotesize]
**Task:**
You will be provided a short video clip (sequence of images) of a human playing Overcooked.
The player may or may not have a defect in their gameplay strategy which is hampering their performance. Your job is to provide corrective feedback to enable the user to understand and fix the defect in their overcooked gameplay.
In addition to the sequence of images, you will also be provided with an analysis of the player's trajectory, which will provide a summary, along with the challenges, successes faced by the player during the trajectory.

**Example Responses (you should follow the same JSON output format):**
{examples}
**Note**
The images are key frames extracted from a video and are presented in the same order as the original footage.
Each player will have at most one defect. The potential defects that the player may have will be one of the following categories:
**{defects}
To help you perform the task, here are some important rules to know about playing Overcooked: 
**{overcooked_rules}

**Now, please look at the following sequence of images carefully.**
{extra}
\end{Verbatim}
  \end{tcolorbox}

  % --- USER (unbreakable) ---
  \begin{tcolorbox}[promptbox, unbreakable,
    title={Corrections Prompt},
    colbacktitle=userc!90!black]
\begin{Verbatim}[fontsize=\footnotesize]
**Task:**
You are an AI assistant tasked with improving a user's performance in Overcooked. 
To help you perform the task, here are some important rules to know about playing Overcooked: 
**{overcooked_rules}
Given a defective seed trajectory, provide corrective actions, that the user should perform next, that align with the user's strategy. 
In addition to the sequence of images, you will also be provided with an analysis of the player's trajectory, which will provide a summary, along with the challenges, successes faced by the player during the trajectory.
Use the accompanying reference trajectory, a non-defective demonstration on a reference map, to infer their strategy. 

**Example Responses (you should follow the same JSON output format):**
{examples}

**Note**
You need to predict the immediate next action the user should take to begin to fix the defect in their behavior. You must choose from one of the following 8 actions: [move forward, move backward, move left, move right, turn left, turn right, interact, no action]
**Now, please look at the following sequence of images carefully.**

{extra}
\end{Verbatim}
  \end{tcolorbox}

  \captionof{figure}{Prompts for GPT-4o baselines with reasoning.}
  \label{fig:GPT-4o prompts with summary trace}
\end{center}

\section{Defects}
\label{sec:defects}
This section provides the full list of defects included in the training-set, held-out-defect dataset and the task-generalization dataset (see Figure~\ref{fig:defects}). 

\begin{center}

  % --- SYSTEM (breakable) ---
  \begin{tcolorbox}[promptbox, breakable,
    title={Defects},
    colbacktitle=systemc!90!black]
\begin{Verbatim}[fontsize=\footnotesize]
## Training Defects
1: "Player does not know that the cooking pot needs to be full for the soup to start cooking.",
2: "Player does not know that you need to interact with dispensers to pick up ingredients.",
3: "Player believes that leaving cooked food on an empty counter means they have delivered the food.",
4: "Player does not know that they need to face the object to successfully interact with any object in the game.",
5: "The player thinks it is best to only serve soup.",
6: "The player thinks tomatoes can be placed on the grill.",
7: "The player thinks onions should be placed directly on the grill.",
8: "Player does not know that cut onions need to be placed on the cooked steak before serving.",
9: "Player mistakenly thinks that the teammate will always walk around them.",
10: "Player does not know that dishes need to be cleaned before being used to pick up food.",
11: "Player does not walk around the visible trip hazards when they are holding an object.",
12: "Player does not remember or keep track of where the invisible trip hazard is." ,
13: "Player attempts to pick up food before it is finished cooking.",
14: "Player is overly cautious about trip hazards when they are holding something, and doesn't go near spaces adjacent to the hazard.",
15: "Player tries to directly hand ingredients they are holding to their teammate.",
16: "Player tries to wash ingredients in the sink.",
"No Defect": "No Defect"

## Heldout defects 
1: "Player does not know that ingredients need to be placed in the cooking ranges to cook.",
2: "The player never serves cooked soup.",
3: "The player thinks it is best to only serve steak.",
4: "The player thinks onions can be cut in the cooking pot.",
5: "The player places ingredients on the counter instead of in the cooking range.",
\end{Verbatim}
  \end{tcolorbox}

  % --- USER (unbreakable) ---
  \begin{tcolorbox}[promptbox, unbreakable,
    title={Defects for Task Generalization Eval},
    colbacktitle=userc!90!black]
\begin{Verbatim}[fontsize=\footnotesize]
## Alternate Task 1
1: "Player forgets to place chopped onions in the cooked soup before serving the soup.",
2: "Player tries to place chopped onions on grilled steak.",

## Alternate Task 2
1: "Player tries to place onions on the chopping board instead of in the cooking pot.",
2: "Player tries to serve cooked tomato and onion soup before picking up chopped steak from the chopping board.",

## Alternate Task 3
1: "Player tries to place three tomatoes into the cooking pot, instead of two tomatoes and an onion.",
2: "Player doesn't know that grilled steak needs to be chopped to make grilled steak skewers.",
\end{Verbatim}
  \end{tcolorbox}

  \captionof{figure}{List of defects used in each dataset}
  \label{fig:defects}
\end{center}

\section{Recipes}
\label{sec:recipes}
Our training dataset uses the original overcooked recipe, i.e. the player can cook either grilled steak and onions or tomato soup. For the task-generalization experiment, the user is prompted to make three alternate recipes: 
\begin{enumerate}
    \item Grilled Steak + Soup and chopped onions
    \item Tomato, Onion and Steak Stew
    \item Tomato and Onion Soup + Grilled Steak Skewers 
\end{enumerate}
The descriptions for each recipe, as utilized in our prompts, are provided in Figure~\ref{fig:recipes})

\begin{center}

  % --- SYSTEM (breakable) ---
  \begin{tcolorbox}[promptbox, breakable,
    title={Recipes},
    colbacktitle=systemc!90!black]
\begin{Verbatim}[fontsize=\footnotesize]
Original Recipe:  Grilled Steak and Onions: The player needs to cook steak on the grill, chop onions on the chopping board, and combine them with the cooked steak. \nTomato Soup: The player needs to place three tomatoes in the cooking pot.

Task-Gen Recipe 1: Grilled Steak:  The player needs to cook steak on the grill and serve it. \n Soup and Chopped Onions: The player needs to make soup by placing three tomatoes in the cooking pot, and add chopped onions after picking up the soup.

Task-Gen Recipe 2: Tomato, Onion and Steak Stew: The player needs to cook steak on the grill and chop it on the chopping board. The player also needs to cook tomato and onion soup, by cooking two tomatoes and one onions in the cooking pot. Finally, the chopped steak should be added to the soup prior to serving.

Task-Gen Recipe 3: Tomato and Onion Soup: The player needs to cook two tomatoes and one onion in the cooking pot. Grilled Steak Skewers: The player needs to cook steak on the grill, then chop it prior to serving.
\end{Verbatim}
  \end{tcolorbox}

  \captionof{figure}{Recipes}
  \label{fig:recipes}
\end{center}

\section{Model Instructions}
\label{sec:model-instructions}
The exact instructions included in the prompt while tuning our model to perform coaching or corrections are provided in Figure~\ref{fig:model-prompts}.

\begin{center}

  % --- SYSTEM (breakable) ---
  \begin{tcolorbox}[promptbox, breakable,
    title={System},
    colbacktitle=systemc!90!black]
\begin{Verbatim}[fontsize=\footnotesize]
You are an AI assistant tasked with providing feedback to a user to improve their performance on the game Overcooked.
<recipe prompt> 

You will be given the user's gameplay trajectory, which includes a specific defect affecting their performance. Your job is to analyze this trajectory and provide corrective feedback to enable the user to understand and fix the defect in their overcooked gameplay.

Defective trajectory: <defective_trajectory>

Corrective Feedback: 
\end{Verbatim}
  \end{tcolorbox}

  % --- USER (unbreakable) ---
  \begin{tcolorbox}[promptbox, unbreakable,
    title={User: },
    colbacktitle=userc!90!black]
\begin{Verbatim}[fontsize=\footnotesize]
You are an AI assistant tasked with improving a user's performance in Overcooked.
<recipe prompt> 

Given a defective seed trajectory, provide corrective actions, that the user should perform next, that align with the user's strategy. Use the accompanying reference trajectory, a non-defective demonstration on a reference map, to infer their strategy.

Reference Trajectory: <reference_trajectory> 

Defective Trajectory: <defective_trajectory>

Corrective Actions: 
\end{Verbatim}
  \end{tcolorbox}

  \captionof{figure}{Model prompts for our model}
  \label{fig:model-prompts}
\end{center}

\section{Visual-QA questions}
\label{sec:vqa-questions}
As explained in Section~\ref{sec:datasets}, we utilize visual QA datasets to improve our model's ability to ground and perform task-generalization. 
This section provides the entire list of questions in the Image/Video/Trajectory-QA datasets (see Figure~\ref{fig:grounding-questions}). 
The Video-QA and Trajectory-QA datasets use the same set of predominantly yes/no questions, whereas the Image-QA dataset features more complex questions requiring a wider range of answers.. 

\begin{center}

  % --- SYSTEM (breakable) ---
  \begin{tcolorbox}[promptbox, breakable,
    title={Image-QA questions},
    colbacktitle=systemc!90!black]
\begin{Verbatim}[fontsize=\footnotesize]
1. 'What is the closest dispenser to the player?'
2. 'Is there a chopped onion on the cutting board?'
3. 'What is the position of the player on the grid?'
4. 'In how many steps can the player reach the delivery location? (Assume that the player cannot walk through hazards if they are holding something)'
5. 'Is there soup cooking in any pot?'
'What is/are the position(s) of the [arg] dispenser on the grid?'
6. 'Are there any items on the counter?'
7. 'Is steak cooking on the grill?'
8. 'Is there cooked steak on the grill?'
9. 'Is there cooked soup in any pot?'
10. 'What is the player holding?'
11. 'How many tomatoes are in the pot(s)?'
12. 'What is the position of the [arg] trip hazard?'
\end{Verbatim}
  \end{tcolorbox}

  % --- USER (unbreakable) ---
  \begin{tcolorbox}[promptbox, unbreakable,
    title={Video/Trajectory-QA questions},
    colbacktitle=userc!90!black]
\begin{Verbatim}[fontsize=\footnotesize]
1. 'What item(s) did the player pickup?'
2. 'Did the player clean a dish?'
3. 'Did the player cut an onion?'
4. 'Did the player trip on a hazard?'
5. 'Did the player place an ingredient in a cooking range?'
6. 'Is the player in a different position from their initial position?'
7. 'Did the player move in a straight line?'
8. 'Did the player successfully deliver food?'
9. 'Is the player closer to the delivery location?'
10. 'Did any food finish cooking?'
\end{Verbatim}
  \end{tcolorbox}

  \captionof{figure}{Questions utilized in each of the visual grounding datasets}
  \label{fig:grounding-questions}
\end{center}

\end{document}